\documentclass[10pt,journal,compsoc]{IEEEtran}
\usepackage[breaklinks=true,bookmarks=false]{hyperref}
\usepackage{comment}
\usepackage{epsfig}
\usepackage{graphicx}
\usepackage{amsmath}
\usepackage{amssymb}
\usepackage{verbatim}
\usepackage{bm}
\usepackage{color}
\usepackage{subfig} 
\usepackage{caption}
\usepackage{booktabs}
\usepackage{subfig}
\usepackage{multirow}
\usepackage{cases}
\ifCLASSOPTIONcompsoc
  \usepackage[nocompress]{cite}
\else
  \usepackage{cite}
\fi
\ifCLASSINFOpdf
\else
\fi
\usepackage{algorithmic}
\usepackage{float}
\usepackage{subfig}

\usepackage{url}

\usepackage[ruled,vlined,linesnumbered,commentsnumbered]{algorithm2e}

\newcommand\gm{{GMTracker}}
\newcommand{\MCG}{\mathcal{G}}
\newcommand{\MCV}{\mathcal{V}}
\newcommand{\MCE}{\mathcal{E}}
\newcommand{\MPI}{\mathbf{\Pi}}
\newcommand{\MA}{\mathbf{A}}

\newcommand{\MBR}{\mathbb{R}}
\newcommand{\MB}{\mathbf{B}}
\newcommand{\Mh}{\mathbf{h}}
\newcommand{\Mm}{\mathbf{m}}
\newcommand{\MCT}{\mathcal{T}}
\newcommand{\MCD}{\mathcal{D}}

\DeclareMathOperator*{\minimize}{minimize}
\DeclareMathOperator*{\diag}{diag}
\DeclareMathOperator*{\subjectto}{subject\;to}
\newcommand{\norm}[1]{\| #1 \|}

\newtheorem{proposition}{Proposition}[section]

\newcommand{\0}{\phantom{0}}
\def\degree{${}^{\circ}$}
\renewcommand{\b}[1]{\textbf{#1}}
\begin{document}
%
\title{Learnable Graph Matching: A Practical Paradigm for Data Association}
\author{Jiawei~He,
        Zehao~Huang,
        Naiyan~Wang,
        and~Zhaoxiang~Zhang, \textit{Senior Member, IEEE}
        \IEEEcompsocitemizethanks{
        \IEEEcompsocthanksitem{
        This work was supported in part by the National Natural Science Foundation of China (No. U21B2042, No. 62072457), and in part by the 2035 Innovation Program of CAS. (Corresponding author: Zhaoxiang Zhang.)}
        \IEEEcompsocthanksitem{
          Jiawei~He and Zhaoxiang~Zhang are with Center for Research on Intelligent Perception and Computing (CRIPAC), State Key Laboratory of Multimodal Artificial Intelligence Systems (MAIS), Institute of Automation, Chinese Academy of Sciences (CASIA), Beijing 100190, China, and also with School of Artificial Intelligence, University of Chinese Academy of Sciences, Beijing 100049, China.  E-mail: \{hejiawei2019, zhaoxiang.zhang\}@ia.ac.cn.}
          \IEEEcompsocthanksitem{Zhaoxiang~Zhang is also with Centre for Artificial Intelligence and Robotics, Hong Kong Institute of Science and Innovation, Chinese Academy of Sciences (HKISI\_CAS), Hong Kong, China.}
          \IEEEcompsocthanksitem{
          Zehao~Huang and Naiyan~Wang are with Tusimple, Beijing 100020,
          China. E-mail: \{zehaohuang18, winsty\}@gmail.com.}}
          }
%
%

\markboth{IEEE TRANSACTIONS ON PATTERN ANALYSIS AND MACHINE INTELLIGENCE,~Vol.~xx, No.~x, xxx~2024}%
{Shell \MakeLowercase{\textit{et al.}}: Bare Demo of IEEEtran.cls for Computer Society Journals}
%



\IEEEtitleabstractindextext{%
\begin{abstract}
  Data association is at the core of many computer vision tasks, e.g., multiple object tracking, image matching, and point cloud registration. however, current data association solutions have some defects: they mostly ignore the intra-view context information; besides, they either train deep association models in an end-to-end way and hardly utilize the advantage of optimization-based assignment methods, or only use an off-the-shelf neural network to extract features. In this paper, we propose a general learnable graph matching method to address these issues. Especially, we model the intra-view relationships as an undirected graph. Then data association turns into a general graph matching problem between graphs. Furthermore, to make optimization end-to-end differentiable, we relax the original graph matching problem into continuous quadratic programming and then incorporate training into a deep graph neural network with KKT conditions and implicit function theorem. In MOT task, our method achieves state-of-the-art performance on several MOT datasets. For image matching, our method outperforms state-of-the-art methods on a popular indoor dataset, ScanNet. For point cloud registration, we also achieve competitive results. Code will be available at \url{https://github.com/jiaweihe1996/GMTracker}.
\end{abstract}

}

\maketitle

\IEEEdisplaynontitleabstractindextext

%
\IEEEpeerreviewmaketitle


%
%
%
%

\IEEEraisesectionheading{\section{Introduction}\label{sec:introduction}}
    \begin{figure*}[t]
        \centering
        \subfloat[Bipartite Matching~\cite{kuhn1955hungarian}]{
            \includegraphics[width=0.46\linewidth]{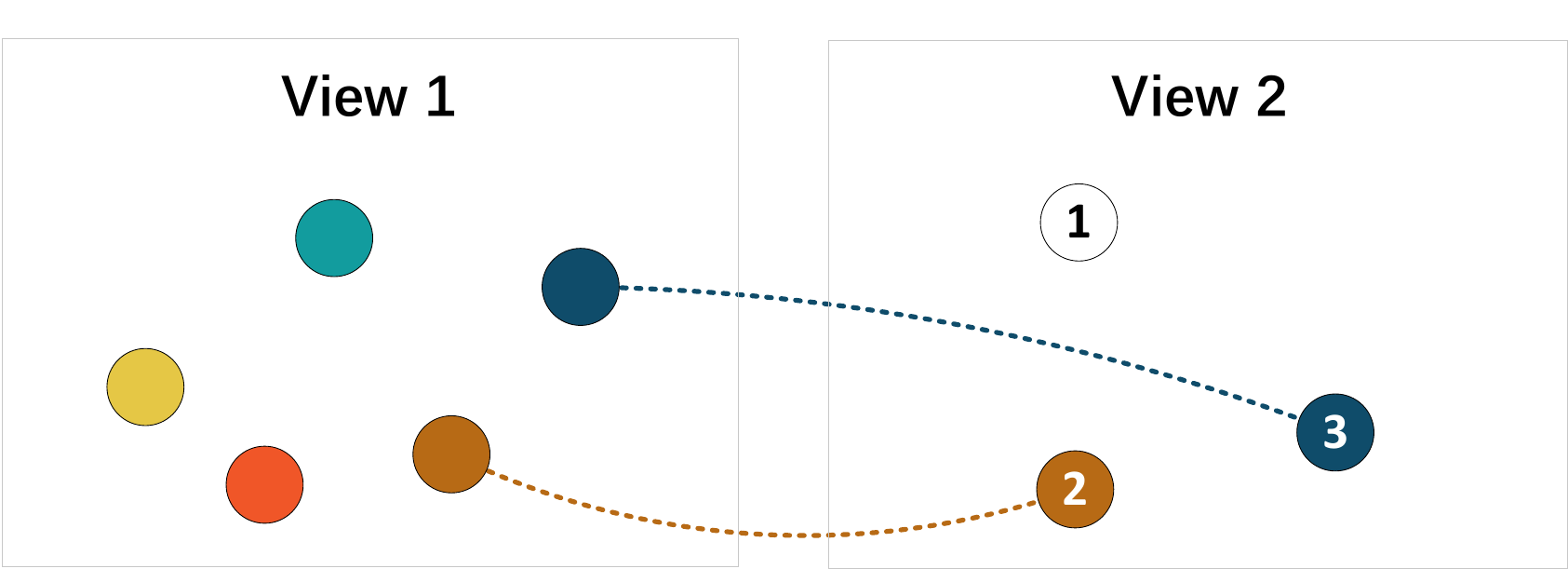}}
            \quad
            \subfloat[Graph Matching]{
            \includegraphics[width=0.46\linewidth]{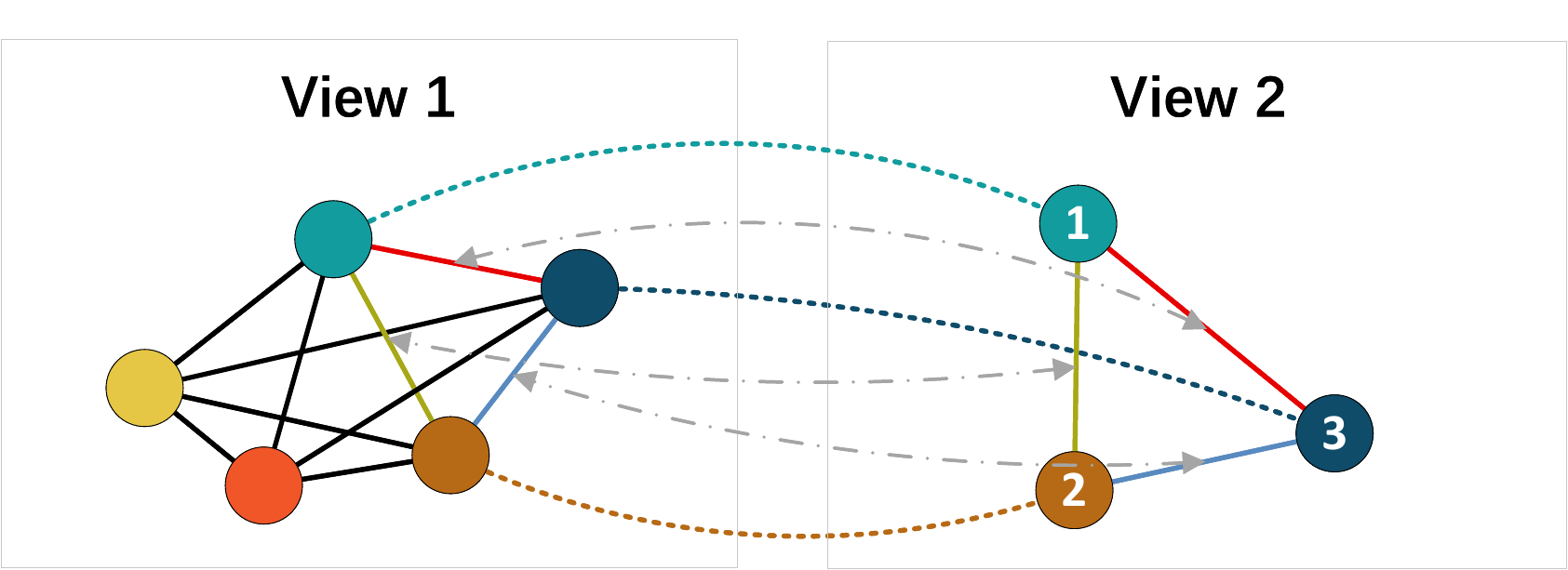}}
            \par
            \subfloat[Multiple Object Tracking]{
            \includegraphics[width=0.31\linewidth]{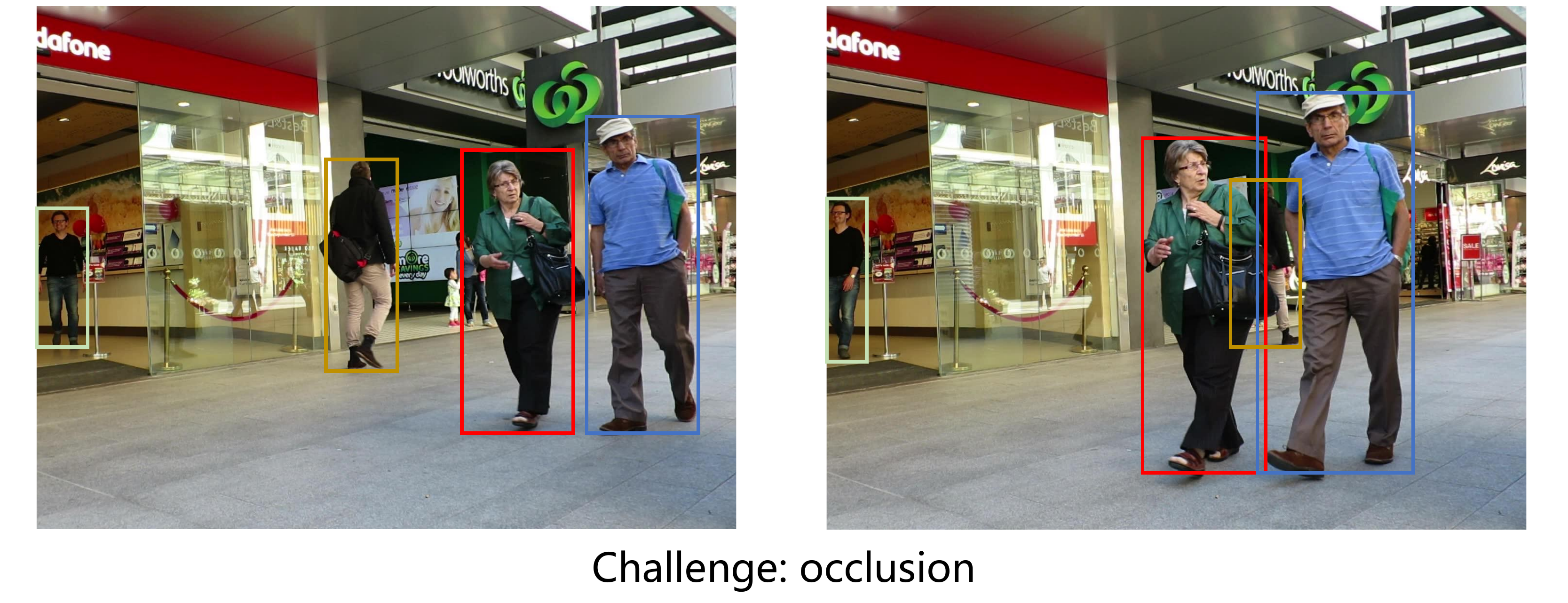}}
            \quad
            \subfloat[Image Matching]{
            \includegraphics[width=0.31\linewidth]{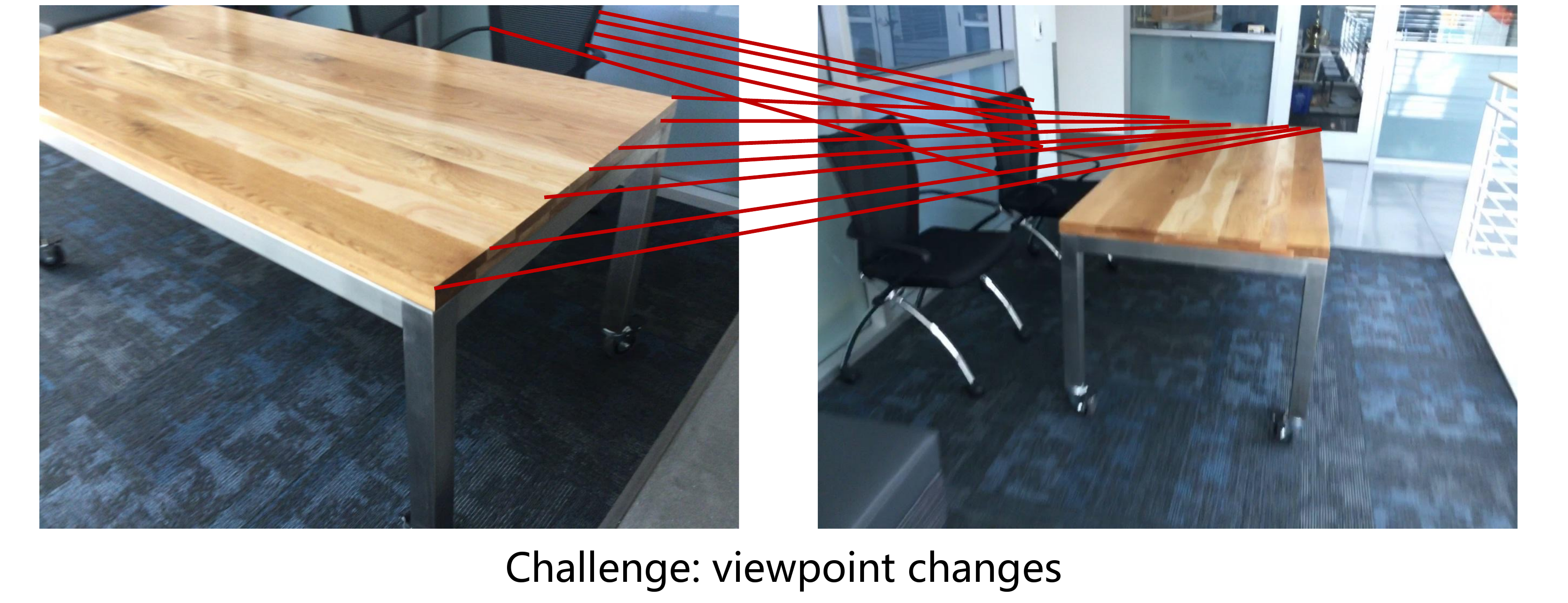}}
            \quad
            \subfloat[Point Cloud Registration]{
            \includegraphics[width=0.31\linewidth]{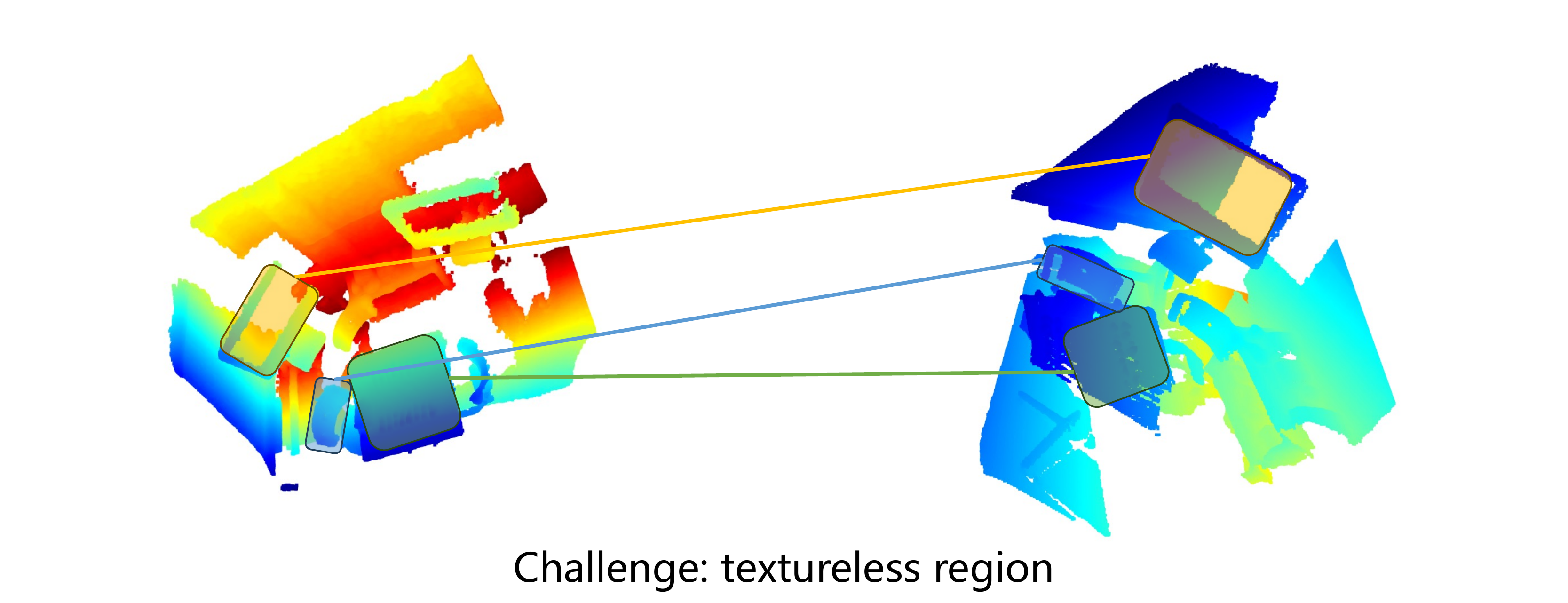}}
        \caption{An illustration of intra-graph relationship used in our graph matching formulation. We utilize the second-order edge to model the pairwise relationship, which is more robust in challenging scenes, such as heavy occluded in MOT task. For example, in view 2, the entity with ID 1 can not be associated with the entity correctly in view 1. However, with graph matching, the pairwise relationship helps data association.}
        \label{fig1}
    \end{figure*}
\IEEEPARstart{D}{ata} association is at the core of many computer vision tasks, for example, instances with the same identity are associated between different frames in \emph{Multiple Object Tracking}, 2D or 3D keypoints are associated between different views for \emph{Image Matching} and \emph{Point Cloud Registration}. These tasks can be described as detecting entities (objects/keypoints/points) in different views, then establishing the correspondences between entities in these views. In this paradigm, the latter process is called \emph{Data Association} and becomes an important part of these tasks. The traditional methods define the data association task as a bipartite matching problem, ignoring the context information in each view, i.e., the pairwise relationship between entities. 
In this paper, we argue that the relationship between the entities within the same view is also crucial for some challenging cases in data association. 
Interestingly, these pairwise relationships within the same view can be represented as edges in a general graph.
To this end, the popular bipartite matching across views can be updated to general graph matching between them. 
To further integrate this novel assignment formulation with powerful feature learning, we first relax the original formulation of graph matching \cite{LawlerMS63, kbqap} to quadratic programming and then derive a differentiable QP layer based on the KKT conditions and the implicit function theorem for the graph matching problem, inspired by the OptNet \cite{amos2017optnet}. Finally, the assignment problem can be learned in synergy with the features. In Fig.~\ref{fig1}, we show how a pairwise relationship is used in our learnable graph matching method.\par 

To reveal the effectiveness and universality of our learnable graph matching method, we apply our method to some important and popular computer vision tasks, \emph{Multiple Object Tracking} (MOT), \emph{Image Matching}, and \emph{Point Cloud Registration}.
MOT is a fundamental computer vision task that aims at associating the same object across successive frames in a video clip. A robust and accurate MOT algorithm is indispensable in broad applications, such as autonomous driving and video surveillance.
The \textit{tracking-by-detection} is currently the dominant paradigm in MOT. This paradigm consists of two steps: (1) obtaining the bounding boxes of objects by detection frame by frame; (2) generating trajectories by associating the same objects between frames. With the rapid development of deep learning-based object detectors, the first step is largely solved by powerful detectors such as~\cite{lin2017feature, lin2017focal}. 
As for the second one, recent MOT work focuses on improving the performance of data association mainly from the following two aspects: (1) formulating the association problem as a combinatorial graph partitioning problem and solving it by advanced optimization techniques \cite{berclaz2011multiple, bewley2016simple, wang2017learning, braso2020learning, xu2020train, hornakova2020lifted}; (2) improving the appearance models by the power of deep learning\cite{kuo2011does, yang2012online, leal2016learning, wojke2017simple}. Although very recently, some work~\cite{zhu2018online, sun2019deep, braso2020learning, xu2020train} trying to unify feature learning and data association into an end-to-end trained neural network, these two directions are almost isolated so that these recent attempts hardly utilize the progress from the combinatorial graph partitioning.\par
In this paper, we propose a learnable graph matching-based online tracker, called GMTracker. We construct the tracklet graph and the detection graph, and put the detections and tracklets on the vertices respectively. The edge features on each graph represent the pairwise relationship between two connected vertices. According to the general design of the learnable graph matching module mentioned above, the learnable vertex and edge features on the two graphs are the input of the differentiable GM layer. The supervision acts on these features constrained by graph matching solutions. So the learning process converges to a more robust and reasonable solution than the traditional approach.\par
However, when the quantity of the objects increases, the graph matching method is not efficient enough. The time cost will become unbearable. So, we speed up the graph matching solver by introducing the gate mechanism, called Gated Search Tree (GST) for MOT. The feasible region is limited and much smaller than the original quadratic programming formulation, so the running time of solving the graph matching problem can be substantially reduced.\par
In the Image Matching task, recently, learning-based methods have been popular. SuperGlue~\cite{sarlin2020superglue} is one of the representative works. Taking the keypoints from traditional methods, e.g., SIFT~\cite{lowe2004distinctive}, or learning-based methods, e.g., SuperPoint~\cite{detone2018superpoint} as the input, SuperGlue proposes transformer-based network to aggregate the keypoint features and matching the corresponding keypoints by a Sinkhorn layer. However, Sinkhorn is only a kind of learnable bipartite matching method, and no intra-frame relationship has been utilized explicitly. Based on SuperGlue~\cite{sarlin2020superglue}, we construct the graph in each frame and replace the Sinkhorn matching module with our differentiable graph matching network, called GMatcher. On the widely used indoor image matching dataset ScanNet~\cite{dai2017scannet}, the result fully embodies the characteristics of fast convergence and good performance of our learnable graph matching method.\par
In summary, our work has the following contributions:
\begin{itemize}
\item Instead of only focusing on data association across views, we emphasize the importance of intra-view relationships. Particularly, we propose to represent the relationships as a general graph, and formulate the data association problem as general graph matching.
\item To solve this challenging assignment problem, and further incorporate it with deep feature learning, we derive a differentiable quadratic programming layer based on the continuous relaxation of the problem, and utilize implicit function theorem and KKT conditions to derive the gradient w.r.t the input features during back-propagation.
\item We design the Gated Search Tree (GST) algorithm, greatly accelerating the process of solving the quadratic assignment problem in data association. Utilizing the new GST algorithm, the association stage is about 21$\times$ faster than the original Quadratic Programming solver. 
\item In the MOT task, we evaluate our proposed {\gm} on the large-scale open benchmark. Our method could remarkably advance state-of-the-art performance in terms of association metrics.
\item In the image matching and point cloud registration tasks, compared with newly reported methods, we achieve SOTA/SOTA comparable performance.
\end{itemize}

\section{Related Work}

\noindent{\bf Data association in MOT.}
The data association step in \textit{tracking-by-detection} paradigm is generally solved by probabilistic filter or combinatorial optimization techniques.
Classical probabilistic approach includes JPDA \cite{bar1990tracking} and MHT \cite{reid1979algorithm}. The advantage of this approach is to keep all the possible candidates for association, and remain the chance to recover from failures. Nevertheless, their costs are prohibitive if no approximation is applied \cite{hamid2015joint, kim2015multiple}. 
For combinatorial optimization, traditional approach include bipartite matching \cite{bewley2016simple}, dynamic programming \cite{fleuret2007multicamera}, min-cost flow \cite{zhang2008global, berclaz2011multiple} and conditional random field \cite{yang2011learning}. Follow-up work tried to adopt more complex optimization methods \cite{zamir2012gmcp, tang2015subgraph}, reduce the computational cost \cite{pirsiavash2011globally, tang2016multi} or promote an online setting from them \cite{choi2015near, wang2016tracklet}.
Early work of deep learning-based association in MOT such as \cite{wojke2017simple,leal2016learning, sadeghian2017tracking,Li_2020_WACV} mostly focus on learning a better appearance model for each object. 
More recently, several work tried to bridge the graph optimization and end-to-end deep learning \cite{jiang2019graph, braso2020learning, xu2020train, Li_2020_WACV, hornakova2020lifted,li2022learning}. \cite{jiang2019graph} adopts Graph Neural Network (GNN) to learn an affinity matrix in a data-driven way. MPNTrack \cite{braso2020learning} introduces a message passing network to learn high-order information between vertices from different frames. \cite{Li_2020_WACV} constructs two graph networks to model appearance and motion features, respectively. 
LifT \cite{hornakova2020lifted} proposes a lifted disjoint path formulation for MOT, which introduces lifted edges to capture long term temporal interactions. \cite{hu2020dual} is the first to formulate the MOT task as a graph matching problem and use dual L1-normalized tensor power iteration method to solve it. Different from \cite{hu2020dual} that directly extracts the features from an off-the-shelf neural network, we propose to guide the feature learning by the optimization problem, which can both enjoy the power of deep feature learning and combinatorial optimization. This joint training manner of representation and optimization problem also eliminate the inconsistencies between the training and inference.

\noindent{\bf Data association in image matching and point cloud registration.} Image matching is a traditional computer vision task, and the main procedures include keypoint detection on the image, keypoint feature extraction, and keypoint matching. Earlier researches mainly focus on the robust keypoint detectors and feature extractors, especially for the local feature descriptor with scale-invariance, rotation-invariance and translation-invariance, such as SIFT~\cite{lowe2004distinctive}, SURF~\cite{bay2006surf} and ORB~\cite{rublee2011orb}. And the matching algorithms are based on the Nearest Neighbor (NN) search with different outlier filtering methods~\cite{lowe2004distinctive,tuytelaars2000wide,sattler2009scramsac}. In the era of deep learning, the CNN-based feature detectors and extractors have emerged, such as D2-net~\cite{dusmanu2019d2} and SuperPoint~\cite{detone2018superpoint}. However, the exploration of end-to-end learnable matching algorithms has just begun. SuperGlue~\cite{sarlin2020superglue} utilizes the attentional graph neural network to model the long-range relationship in the image, and the matching module is designed as a differentiable Sinkhorn layer, which is the approximation of graph matching without explicitly modeling the edge in the graph. LoFTR~\cite{sun2021loftr} is a kind of dense matching algorithm, from patch matching to pixel matching in coarse-to-fine style. 
As for point cloud registration, there are two kinds of methods: correspondence-based methods and direct registration methods. The correspondence-based methods~\cite{deng2018ppfnet, choy2019fully,qin2022geometric} extract correspondences between two point clouds and solve the transformation matrix with pose estimators, e.g. SVD, RANSAC, LGR. Direct registration methods~\cite{wang2019prnet, huang2020feature, fu2021robust} use an end-to-end neural network to regress the transformation matrix.\par
\noindent{\bf Graph matching and combinatorial optimization.}
Pairwise graph matching, or more generally Quadratic Assignment Problem (QAP), has wide applications in various computer vision tasks \cite{vento2013graph}.
Compared with the linear assignment problem that only considers vertex-to-vertex relationship, pairwise graph matching also considers the second-order edge-to-edge relationship in graphs. 
The second-order relationship makes matching more robust. However, as shown in \cite{hartmanis1982computers}, this problem is an NP-hard problem. 
There is no polynomial solver like Hungarian algorithm~\cite{kuhn1955hungarian} for the linear assignment problem. In the past decades, many work engages in making the problem tractable by relaxing the original QAP problem~\cite{leordeanu2005spectral,schellewald2005probabilistic,torr2003solving}. Lagrangian decomposition~\cite{swoboda2017study} and factorized graph matching~\cite{facgm} are two representative ones. 
To incorporate graph matching into deep learning, one stream of work is to treat the assignment problem as a supervised learning problem directly, and use the data fitting power of deep learning to learn the projection from input graphs to output assignment directly ~\cite{wang2019learning,Yu2020Learning}.
Another more theoretically rigorous is to relax the problem to a convex optimization problem first, and then utilize the KKT condition and implicit function theorem to derive the gradients w.r.t all variables at the optimal solution ~\cite{barratt2018differentiability}. Inspired by it, in this paper, we propose a learnable graph matching layer to solve the challenging graph matching problem in data association.




\section{Graph Matching Formulation for Data Association}
\label{sec:relax}
In this section, we will formulate data association problem as a graph matching problem. Instead of solving the original Quadratic Assignment Problem (QAP), we relax the graph matching formulation as a convex quadratic programming (QP) and extend the formulation from the edge weights to the edge features. The relaxation facilitates the differentiable and joint learning of feature representation and combinatorial optimization.
\subsection{Basic Graph Matching Formulation for Data Association}
We define the aim of data association is to match the vertices in graph $\mathcal{G}_1$ and $\mathcal{G}_2$ constructd in view 1 and view 2 respectively. So, it can be seen as a graph matching problem, which is to maximize the similarities between the matched vertices and corresponding edges connected by these vertices.
As defined in \cite{LawlerMS63}, the graph matching problem is a Quadratic Assignment Problem (QAP) . A practical mathematical form is named \emph{Koopmans-Beckmann's} QAP \cite{kbqap}:
\begin{equation}
\label{equ:KBQAP}
\begin{aligned}
& \underset{\MPI}{\text{maximize}}
&& \mathcal{J}(\MPI)=\text{tr}(\MA_1\MPI\MA_2\MPI^\top)+\text{tr}(\MB^\top\MPI),  \\
& \text {s.t.}
&& \MPI\mathbf{1}_{n}= \mathbf{1}_{n}, \MPI^\top\mathbf{1}_{n}= \mathbf{1}_{n},
\end{aligned}
\end{equation}
where $\MPI\in\{0,1\}^{n\times{n}}$ is a permutation matrix that denotes the matching between the vertices of two graphs, $\MA_1\in\MBR^{n\times n}$, $\MA_2\in\MBR^{n\times n}$ are the weighted adjacency matrices of graph $\mathcal{G}_1$ and $\mathcal{G}_2$ respectively, and $\MB\in\MBR^{n\times n}$ is the vertex affinity matrix between $\mathcal{G}_1$ and $\mathcal{G}_2$. $\mathbf{1}_{n}$ denotes an n-dimensional vector with all values to be 1.

\subsection{Reformulation and Convex Relaxation}
\label{sec:qp}
For \emph{Koopmans-Beckmann's} QAP, as $\MPI$ is a permutation matrix, i.e., $\MPI^\top\MPI=\MPI\MPI^\top=\mathbf{I}$. Following~\cite{facgm}, Eq.~\ref{equ:KBQAP} can be rewritten as
\begin{equation}
\begin{aligned}
\mathbf{\Pi}^*
&=\underset{\mathbf{\Pi}}{\arg\min} \ \frac{1}{2}||\mathbf{A_1}\mathbf{\Pi}-\mathbf{\Pi}\mathbf{A_2}||_F^2-\text{tr}(\mathbf{B}^\top\mathbf{\Pi}).
\end{aligned}
\label{K-B}
\end{equation}
This formulation is more intuitive than that in Eq.~\ref{equ:KBQAP}. For two vertices $i, i' \in \MCG_1$  
and their corresponding vertices $j, j' \in \MCG_2$, the first term in Eq.~\ref{K-B} denotes the difference of the weight of edge $(i, i')$ and $(j, j')$, and the second term denotes the vertex affinities between $i$ and $j$. Then the goal of the optimization is to maximize the vertex affinities between all matched vertices, and minimize the difference of edge weights between all matched edges.
 
It can be proven that the convex hull of the permutation matrix lies in the space of the doubly-stochastic matrix. So, as shown in \cite{aflalo2015convex}, the QAP (Eq.~\ref{K-B}) can be relaxed to its tightest convex relaxation by only constraining the permutation matrix $\mathbf{\Pi}$ to be a double stochastic matrix $\mathbf{X}$, formed as the following QP problem:
\begin{equation}
\mathbf{X}^*=\underset{\mathbf{X}\in\mathcal{D}}{\arg\min} \ \frac{1}{2}||\mathbf{A_1}\mathbf{X}-\mathbf{X}\mathbf{A_2}||_F^2-\text{tr}(\mathbf{B}^\top\mathbf{X}),
\label{QP}
\end{equation}
where $\mathcal{D}=\{\mathbf{X}:\mathbf{X}\mathbf{1}_n= \mathbf{1}_n, \mathbf{X}^\top\mathbf{1}_n=\mathbf{1}_n,\mathbf{X}\geq\mathbf{0}\}$.
\section{Graph Matching for MOT}
In this section, we introduce the problem definition of MOT and our graph matching formulation for the data association in MOT task.
\subsection{Detection and Tracklet Graphs Construction}
\label{sec:construct}
As an online tracker, we track objects frame by frame. 
In frame $t$, we define $\MCD^t=\{D_1^t, D_2^t,\cdots, D_{n_d}^t\}$ as the set of detections in current frame and $\MCT^t=\{T_1^t, T_2^t, \cdots, T_{n_t}^t\}$ as the set of tracklets obtained from past frames. $n_d$ and $n_t$ denote the number of detected objects and tracklet candidates. A detection is represented by a triple $D_p^t=(\mathbf{I}_p^t, \mathbf{g}_p^t, t)$, where $\mathbf{I}_p^t$ contains the image pixels in the detected area, $\mathbf{g}_p^t=(x_p^t,y_p^t,w_p^t,h_p^t)$ is a geometric vector including the central location and size of the detection bounding box. Each tracklet contains a series of detected objects with the same tracklet id. With a bit abuse of notations, the generation of $T_{id}^t$ can be represented as $T_{id}^{t} \gets T_{id}^{t-1} \cup \{D^{t-1}_{(id)}\}$, which means we add $D^{t-1}_{(id)}$ to the tracklet $T_{id}^{t-1}$.

Then we define the detection graph in frame $t$ as $\MCG_D^t=(\MCV_D^t, \MCE_D^t)$ and the tracklet graph up to the frame $t$ as $\MCG_T^t=(\MCV_T^t, \MCE_T^t)$. Each vertex $i \in \MCV_D^t$ and vertex $j\in \MCV_T^t$ represents the detection $D_i^t$ and the tracklet $T_j^t$, respectively. The $e_u=(i,i')$ is the edge in $\MCE_D^t$ and $e_v=(j,j')$ is the edge in $\MCE_T^t$. Both of these two graphs are complete graphs. Then the data association in frame $t$ can be formulated as a graph matching problem between $\MCG_D^t$ and $\MCG_T^t$. For simplicity, we will ignore $t$ in the following sections.
\subsection{From Edge Weights to Edge Features}
In the general formulation of graph matching, the element $a_{i,i'}$ in the weighted adjacency matrix $\MA \in\MBR^{n\times n}$ is a scalar denoting the weight on the edge $(i, i')$. To facilitate the application in our MOT problem, we expand the relaxed QP formulation by using an \emph{$l_2$-normalized} edge feature $\mathbf{h}_{i,i'} \in \MBR^d$ instead of the scalar-formed edge weight $a_{i,i'}$ in $\mathbf{A}$. We build a weighted adjacency tensor $\mathbf{H} \in \MBR^{d \times n \times n}$ where $\mathbf{H}^{\cdot, i,i'}$ = $\mathbf{h}_{i, i'}$, i.e., we consider the each dimension of $\mathbf{h}_{i,i'}$ as the element $a_{i,i'}$ in $\mathbf{A}$ and concatenate them along channel dimension. The $\mathbf{H_D}$ and $\mathbf{H_T}$ are the weighted adjacency tensors for $\MCG_D$ and $\MCG_T$, respectively. Then the optimization objective in Eq.~\ref{K-B} can be further expanded to consider the $l_2$ distance between two corresponding \emph{n-d} edge features other than the scalar differences:
\begin{equation}
\begin{aligned}
\mathbf{\Pi}^*
&=\underset{\mathbf{\Pi}}{\arg\min} \ \sum_{c=1}^{d}\frac{1}{2}||\mathbf{H}_D^c\mathbf{\Pi}-\mathbf{\Pi}\mathbf{H}_T^c||_F^2-\text{tr}(\mathbf{B}^\top\mathbf{\Pi}) \\
&=\underset{\mathbf{\Pi}}{\arg\min} \ \sum_{i=1}^{n}\sum_{i'=1}^{n}\sum_{j=1}^{n}\sum_{j'=1}^{n}  \frac{1}{2}||\mathbf{h}_{ii'}\pi_{ij}-\mathbf{h}_{jj'}\pi_{i'j'}||_2^2 \\
&\ \ \ \ -\text{tr}(\mathbf{B}^\top\mathbf{\Pi}) \\
&=\underset{\mathbf{\Pi}}{\arg\min} \ \sum_{i=1}^{n}\sum_{i'=1}^{n}\sum_{j=1}^{n}\sum_{j'=1}^{n}  \frac{1}{2}(\pi_{ij}^2-2\pi_{ij}\pi_{i'j'}\mathbf{h}_{ii'}^\top\mathbf{h}_{jj'} \\
&\ \ \ \ +\pi_{i'j'}^2)-\text{tr}(\mathbf{B}^\top\mathbf{\Pi}),
\end{aligned}
\label{before_edge}
\end{equation}
where $n$ is the number of vertices in graph $\mathcal{G}_D$ and $\mathcal{G}_T$, the subscript $i$ and $i'$ are the vertices in graph $\mathcal{G}_D$ and $j$ and $j'$ are in graph $\mathcal{G}_T$. We reformulate Eq.~\ref{before_edge} as: 
\begin{equation}
\bm{\pi}^*=\underset{\bm{\pi}}{\arg\min} \ 
\bm{\pi}^\top((n-1)^2\mathbf{I}-\mathbf{M})\bm{\pi}-\mathbf{b}^\top\bm{\pi},
\label{edge}
\end{equation}
where $\bm{\pi}=\text{vec}(\MPI)$, $\mathbf{b}=\text{vec}(\MB)$ and $\mathbf{M}\in \MBR^{n^2\times n^2}$ is the symmetric quadratic affinity matrix between all the possible edges in two graphs.

Following the relaxation in Section \ref{sec:qp}, the formulation Eq.~\ref{edge} using edge features can be relaxed to a QP:\\
\begin{equation}
\begin{aligned}
\mathbf{x}^*
&=\underset{\mathbf{x}\in\mathcal{D^{'}}}{\arg\min} \ 
\mathbf{x}^\top((n-1)^2\mathbf{I}-\mathbf{M})\mathbf{x}-\mathbf{b}^\top\mathbf{x}, \\
\end{aligned}
\label{finalQP}
\end{equation}
where $ \mathcal{D}^{'}=\{\mathbf{x}:\mathbf{R}\mathbf{x}=\mathbf{1},\mathbf{U}\mathbf{x}\leq \mathbf{1},\mathbf{x}\geq\mathbf{0},\mathbf{R}=\mathbf{1}_{n_2}^\top\otimes{\mathbf{I}_{n_1}},\mathbf{U}=\mathbf{I}_{n_2}^\top\otimes{\mathbf{1}_{n_1}}\}$, $\otimes$ denotes Kronecker product.
\begin{figure}
          \centering
           \includegraphics[width=\linewidth]{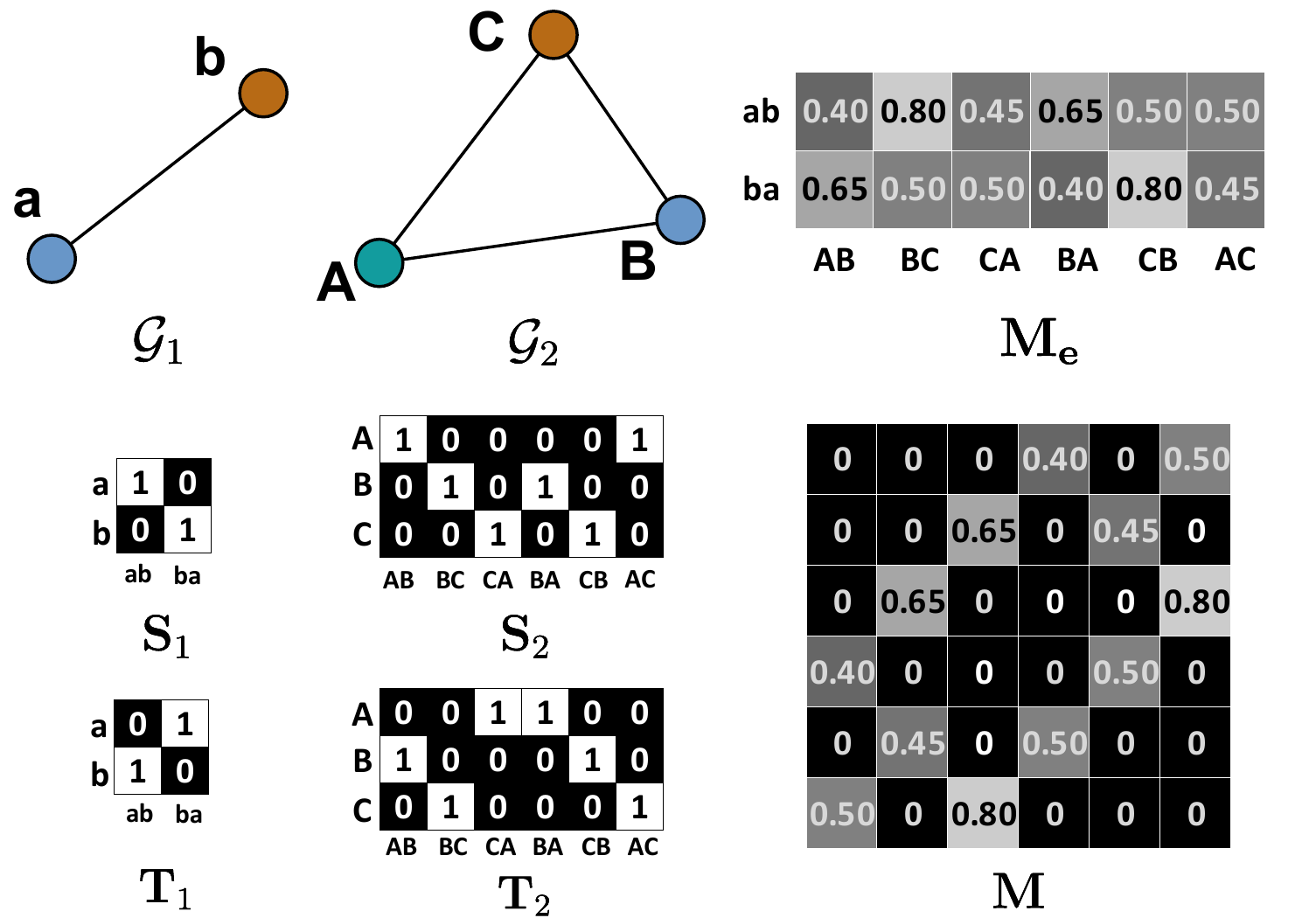}
             \caption{An example of the derivation from edge affinity matrix $\mathbf{M_e}$ to quadratic affinity matrix $\mathbf{M}$.}
             \label{fig:me2m}
  \end{figure}    

 In the implementation, we first compute the cosine similarity between the edges in $\mathcal{G}_D$ and $\mathcal{G}_T$ to construct the matrix $\mathbf{M_e}\in\MBR^{|\mathcal{E}_D|\times |\mathcal{E}_T|}$. 
 The element of the matrix $\mathbf{M_e}$ is the cosine similarity between edge features $\mathbf{h}_{i,i'}$ and $\mathbf{h}_{j,j'}$ in two graphs:
\begin{align}
  \mathbf{M}_e^{u,v}=\mathbf{h}_{i,i'}^\top\mathbf{h}_{j,j'},
  \label{eq:e2e}
\end{align}
where $e_u=(i,{i'})$ is the edge in $\mathcal{G}_D$ and $e_v=(j,{j'})$ is the edge in $\mathcal{G}_T$. 
 
 And following \cite{zanfir2018deep}, we map each element of matrix $\mathbf{M_e}$ to the \emph{symmetric} quadratic affinity matrix $\mathbf{M}$:
\begin{align}
  \mathbf{M}=(\mathbf{S_D}\otimes\mathbf{S_T})\text{diag}(\text{vec}(\mathbf{M_e}))(\mathbf{T_D}\otimes\mathbf{T_T})^\top,
  \label{eq:me2m}
\end{align}
where $\text{diag}(\cdot)$ means constructing a diagonal matrix by the given vector, $\mathbf{S_D} \in \{0,1\}^{|\MCV_D| \times |\MCE_D|}$ and $\mathbf{S_T} \in \{0,1\}^{|\MCV_T| \times |\MCE_T|}$, whose elements are an indicator function:
\begin{equation}
  \mathbb{I}_s(i,u):= \begin{cases}
    1 & \text{if $i$ is the start vertex of edge $e_u$}, \\
    0 & \text{if $i$ is not the start vertex of edge $e_u$},
  \end{cases}
\end{equation}
$\mathbf{T_D} \in \{0,1\}^{|\MCV_D| \times |\MCE_D|}$ and $\mathbf{T_T} \in \{0,1\}^{|\MCV_T| \times |\MCE_T|}$, whose elements are another indicator function:
\begin{equation}
  \mathbb{I}_t(i',u):= \begin{cases}
    1 & \text{if $i'$ is the end vertex of edge $e_u$}, \\
    0 & \text{if $i'$ is not the end vertex of edge $e_u$}.
  \end{cases}
\end{equation}
An example of the derivation from $\mathbf{M_e}$ to $\mathbf{M}$ is illustrated in Fig.~\ref{fig:me2m}. 

Besides, each element in the vertex affinity matrix $\mathbf{B}$ is the cosine similarities between feature $\mathbf{h}_i$ on vertex $i \in \MCV_D$ and feature $\mathbf{h}_j$ on vertex $j\in \MCV_T$:
\begin{align}
  \mathbf{B}_{{i,j}}=\mathbf{h}_{i}^\top\mathbf{h}_{j}
  \label{eq:n2n}
\end{align}
\par
\begin{figure*}[ht]
          \centering
           \includegraphics[width=\linewidth]{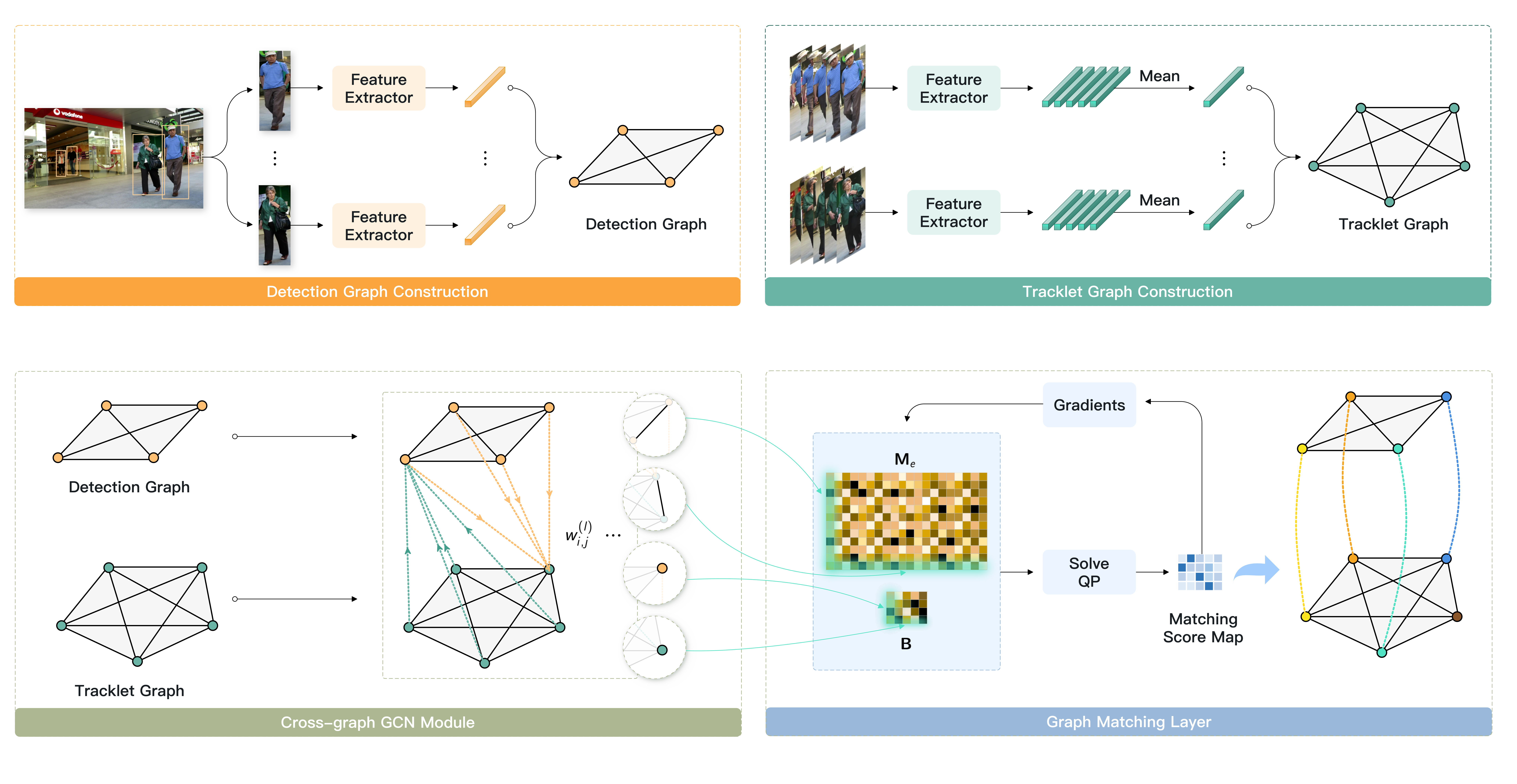}
             \caption{Overview of our GMTracker method. We first extract features from detections and construct the detection graph using these features. The tracklet graph construction step is similar to the detection graph, but we average the features in a tracklet. Then the cross-graph GCN is adopted to enhance the features. The weight $w_{i,j}$ is from the feature similarity and geometric information. The core of our method is the differentiable graph matching layer built as a QP layer from the formulation in Eq. \ref{finalQP}. The $\mathbf{M}_e$ and $\mathbf{B}$ in the graph matching layer denote the edge affinity matrix from Eq. \ref{eq:e2e} and the vertex affinity matrix from Eq. \ref{eq:n2n} respectively.}
             \label{pipeline}
  \end{figure*}  
\section{Graph Matching Network and GMTracker}
In this section, we will describe the details of our Graph Matching Network and our {\gm}. As shown in Fig.~\ref{pipeline}, the pipeline of our Graph Matching Network consists of three parts: (1) feature encoding in detection and tracklet graphs; (2) feature enhancement by cross-graph Graph Convolutional Network (GCN) and (3) differentiable graph matching layer. We will describe these three parts step by step and show how we integrate them into a tracker ({\gm}) in the following.

\subsection{Feature Encoding in Two Graphs}

We utilize a pre-trained ReIDentification (ReID) network followed by a multi-layer perceptron (MLP) to generate the appearance feature $\mathbf{a}_D^{i}$ for each detection $D_i$.
The appearance feature $\mathbf{a}_T^{j}$ of the tracklet $T_{j}$ is obtained by averaging all the appearance features of detections before.\par

\subsection{Cross-Graph GCN}
\label{sec:gcn}
 Similar to \cite{braso2020learning, ma2019deep, Weng2020_GNN3DMOT}, we only adopt a GCN module between the graph $\MCG_D$ and graph $\MCG_T$ to enhance the feature, and thus it is called Cross-Graph GCN. 

 The initial vertex features on detection graph and tracklet graph are the appearance features on the vertices, i.e., let $\Mh_i^{(0)}=\mathbf{a}_D^{i}$ and $\Mh_j^{(0)}=\mathbf{a}_T^{j}$.
Let $\Mh_i^{(l)}$ and $\Mh_j^{(l)}$ be the feature of vertex $i \in \MCG_D$ and vertex $j \in \MCG_T$ in the $l$-th propagation, respectively. 
We define the aggregation weight coefficient $w_{i, j}^{(l)}$ in GCN as the appearance and geometric similarity between vertex $i$ and vertex $j$:
\begin{equation}
w_{i,j}^{(l)} = \cos(\Mh_i^{(l)}, \Mh_j^{(l)}) + {\rm{IoU}}(\mathbf{g}_i, \mathbf{g}_j)
\end{equation}
where $\cos(\cdot, \cdot)$ means the cosine similarity between input features and $\rm{IoU(\cdot, \cdot)}$ denotes the Intersection over Union of two bounding boxes. For a detection vertex $i$, $\mathbf{g}_i$ is the corresponding detection bounding box defined in Section \ref{sec:construct}. As for a tracklet vertex $j$, we estimate the bounding box $\mathbf{g}_j$ in current frame $t$ by Kalman Filter \cite{kalman1960new} motion model with a constant velocity. Note that we only consider the appearance feature similarity in weight $w_{i,j}$ when the camera moves, since the motion model cannot predict reliable future positions in these complicated scenes.

We use summation as the aggregation function, i.e., $\Mm_{i}^{(l)} = \sum_{j \in \MCG_T} w_{i,j}^{(l)}\Mh_j^{(l)}$ and the vertex features are updated by:
\begin{equation}
\Mh_i^{(l + 1)} = {\rm{MLP}}(\Mh_i^{(l)} + \frac{\norm{\Mh_i^{(l)}}_2\Mm_{i}^{(l)}}{\norm{\Mm_{i}^{(l)}}_2}),
\end{equation}
where we adopt message normalization proposed in \cite{li2020deepergcn} to stabilize the training.

We apply $l_2$ normalization to the final features after cross-graph GCN and denote it as $\Mh_i$. Then we use $\Mh_i$ as the feature of vertex $i$ in graph $\MCG_D$, and construct the edge feature for edge $(i, i')$ with $\Mh_{i,i'} = l_2([\Mh_i, \Mh_{i'}])$, where $[\cdot]$ denotes concatenation operation. The similar operation is also applied to the tracklet graph $\MCG_T$. In our implementation, we only apply GCN once.

\subsection{Differentiable Graph Matching Layer}
\label{sec:diffgm}
After enhancing the vertex features and constructing the edge features on graph $\MCG_D$ and $\MCG_T$, we meet the core component of our method: the differentiable graph matching layer. By optimizing the QP in Eq.~\ref{finalQP} from quadratic affinity matrix $\mathbf{M}$ and vertex affinity matrix $\mathbf{B}$, we can derive the optimal matching score vector $\mathbf{x}$ and reshape it back to the shape $n_d \times n_t$ to get the matching score map $\mathbf{X}$.

Since we finally formulate the graph matching problem as a QP, we can construct the graph matching module as a differentiable QP layer in our neural network. Since KKT conditions are the necessary and sufficient conditions for the optimal solution $\mathbf{x}^*$ and its dual variables, we could derive the gradient in backward pass of our graph matching layer based on the KKT conditions and implicit function theorem, which is inspired by OptNet~\cite{amos2017optnet}.
In our implementation, we adopt the qpth library~\cite{amos2017optnet} to build the graph matching module.
In the inference stage, to reduce the computational cost and accelerate the algorithm, we solve the QP using the CVXPY library~\cite{diamond2016cvxpy} only for forward operation.\par
For training, we use weighted binary cross entropy Loss: 
\begin{equation}
  \begin{aligned}
    \mathcal{L}=\frac{-1}{n_dn_t}\sum_{i=1}^{n_d}\sum_{j=1}^{n_t}\ ky_{i,j}\log(\hat{y}_{i,j})+(1-y_{i,j})\log(1-\hat{y}_{i,j}),
\end{aligned}
\end{equation}
where $\hat y_{i,j}$ denotes the matching score between detection $D_i$ and tracklet $T_j$, and $y_{i,j}$ is the ground truth indicating whether the object belongs to the tracklet. $k=(n_t-1)$ is the weight to balance the loss between positive and negative samples. Besides, due to our QP formulation of graph matching, the distribution of matching score map $\mathbf{X}$ is relatively smooth. We adopt softmax function with temperature $\tau$ to sharpen the distribution of scores before calculating the loss:
\begin{equation}
\hat{y}_{i,j} = {\rm Softmax}(x_{i,j},\tau)=\frac{e^{x_{i,j}/\tau}}{\sum_{j=1}^{n_t}e^{x_{i,j}/\tau}},
\label{eq:softmax}
\end{equation}
where $x_{i,j}$ is the original matching score in score map $\mathbf{X}$.

\subsection{Gradients of the Graph Matching Layer}
The gradients of the graph matching layer we need for backward can be derived from the KKT conditions with the help of the implicit function theorem. Here, we show the details of deriving the gradients of a standard QP optimization. 
\par
For a quadratic programming (QP), the standard formulation is as
\begin{equation}
    \begin{aligned}
    & \minimize_x
    & & \frac{1}{2}x^\top Q(\theta) x + q(\theta)^\top x \\
    & \subjectto
    && G(\theta)x \leq h(\theta) \\
    &&& A(\theta)x=b(\theta).
    \end{aligned}
    \end{equation}
So the Lagrangian is given by 
\begin{equation}
    L(x,\nu,\lambda)=\frac{1}{2}x^\top Qx+\lambda^\top (Gx-h)+q^\top x+\nu^\top (Ax-b),
\end{equation}
where, $\nu$ and $\lambda$ are the dual variables.\\
The $(x^*,\lambda^*,\nu^*)$ are the optimal solution if and only if they satisfy the KKT conditions:
\begin{equation}
    \begin{split}
    \nabla_x L(x^*,\lambda^*,\nu^*) &= 0 \\
    Qx^* +q+A^\top \nu^*+G^\top \lambda^* &= 0 \\
    Ax^*-b &= 0 \\
    \diag (\lambda^*)(Gx^*-h) &= 0 \\
    Gx^* -h &\leq 0 \\
    \lambda^*&\geq 0.
    \end{split}
    \end{equation}
We define the function
\begin{equation}
    g(x,\lambda,\nu,\theta) = \begin{bmatrix}
    \nabla_x L({x},\lambda,\nu,\theta) \\
    \diag(\lambda)\lambda^\top (G(\theta)x-h(\theta)) \\
    A(\theta)x-b(\theta)
    \end{bmatrix},
    \end{equation}
and the optimal solution $x^*, \lambda^*, \nu^*$ satisfy the equation $g(x^*, \lambda^*, \nu^*,\theta)=0$. \\
According to the implicit function theorem, as proven in \cite{barratt2018differentiability}, the gradients where the primal variable $x$ and the dual variables $\nu$ and $\lambda$ are the optimal solution, can be formulated as 
    \begin{equation}
        J_\theta x^* = -J_x g(x^*,\lambda^*,\nu^*,\theta)^{-1} J_\theta g(x^*,\lambda^*,\nu^*,\theta),
    \label{eq:jacobian}
    \end{equation}
where, $J_x g(x^*,\lambda^*,\nu^*,\theta)$ and $J_\theta g(x^*,\lambda^*,\nu^*,\theta)$ are the Jacobian matrices. Each element of them is the partial derivative of function $g$ with respect to variable $x$ and $\theta$, respectively.
\subsection{Inference Details}
\label{sec:tarcker}
Due to the continuous relaxation, the output of the QP layer may not be binary. To get a valid assignment, we use the greedy rounding strategy to generate the final permutation matrix from the predicted matching score map, i.e., we match the detection with the tracklet with the maximum score. 
After matching, like DeepSORT \cite{wojke2017simple}, we need to handle the born and death of tracklets. We keep the matching between detection and tracklet only if it satisfies all the following constraints: 1) The appearance similarity between detection and tracklet is above the threshold $\sigma$. 2) The detection is not far away from the tracklet. We set a threshold $\kappa$ as the Mahalanobis distance between the predicted distribution of the tracklet bounding box by the motion model and the detection bounding box in pixel coordinates, called the motion gate. 3) The detection bounding box overlaps with the position of tracklet predicted by the motion model. The constraints above can be written as
    \begin{equation}
        \left\{
    \begin{aligned}
    &\mathbf{B}_{i,j}>\sigma,\\ 
    &\mathtt{KF}_{i,j}>\kappa,\\ 
    &\mathtt{iou}_{i,j}>0,
\end{aligned}
        \right.
    \label{eq:cons}
\end{equation}
where, $(i.j) \in (\mathcal{V}_D,\mathcal{V}_T)$.

Here, besides the Kalman Filter adopted to estimate the geometric information in Section \ref{sec:gcn}, we apply an Enhanced Correlation Coefficient (ECC) \cite{ecc} in our motion model additionally to compensate the camera motion. Besides, we apply the IoU association between the filtered detections and the unmatched tracklets by the Hungarian algorithm to compensate for some incorrect filtering.
 Then the remaining detections are considered as a new tracklet. We delete a tracklet if it has not been updated since $\delta$ frames ago, called \emph{max age}. 
 \begin{figure}[h] 
    \centering
    \includegraphics[width=0.7\linewidth]{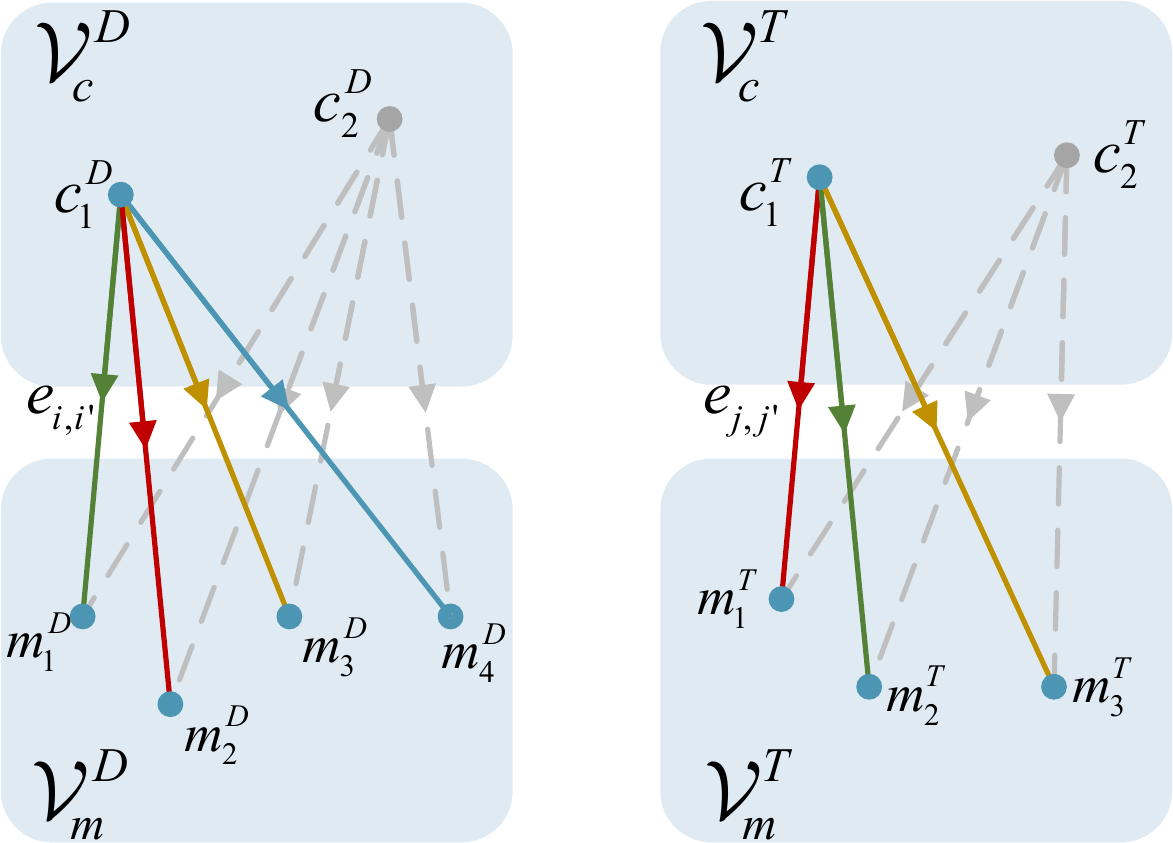}
    \caption{An illustration of edge matching. Here, for matched pair $(c_1^D,c_1^T)$ in $\bm{\pi}_c$, we find best matching between edge $e_{1,i'}$ and $e_{1,j'}$, drawn in the same color.}
    \vspace{-10pt}
    \label{fig:edgematch}
\end{figure}
\subsection{GST: A Practical Algorithm for Quadratic Assignment}
However, due to the process of solving the quadratic programming, the inference speed is relatively slow compared with other mainstream MOT algorithms. To speed up, we design the gated search tree (GST) algorithm. Utilizing constraints Eq.~\ref{eq:cons}, the feasible region is limited much smaller than the original quadratic programming, which greatly accelerates the process of solving the quadratic assignment problem.\par
The GST algorithm (Alg.~\ref{alg:gst}) contains three main steps. Firstly, we construct a bipartite graph $\mathcal{G}=(\mathcal{V}_D,\mathcal{V}_T,\mathcal{E}_{DT})$, in which the edges are only between tracklets and detections meeting the constraints, i.e., $\mathcal{E}_{DT}=\{(i,j)|i\in \mathcal{V}_D,j\in \mathcal{V}_T, \mathbf{B}_{i,j}>\sigma, \mathtt{KF}_{i,j}>\kappa,
 \mathtt{iou}_{i,j}>0\}$. Secondly, we use depth-first search to find all the connected components $\{\mathcal{G}_c\}=\{\mathcal{G}_c=(\mathcal{V}^D_c,\mathcal{V}^T_c,\mathcal{E}^{DT}_c)|c=(1,2,\cdots,k),|\mathcal{V}^D_1|+|\mathcal{V}^T_1|\geq|\mathcal{V}^D_2|+|\mathcal{V}^T_2|\geq\cdots\geq|\mathcal{V}^D_k|+|\mathcal{V}^T_k|\}$.
 Last, we calculate the matching cost $\mathcal{L}(\bm{\pi}_c)$ for all matching candidates in each independent connected component $\mathcal{G}_c$ parallelly.\par
The matching cost follows the objective function of the quadratic programming Eq.~\ref{finalQP}. However, as a searching algorithm, the convex relaxation in the objective function shows no advantage. So, the matching cost can be denoted back to the objective function of the original QAP, as
\begin{equation}
    \mathcal{L}(\bm{\pi})=-\bm{\pi}^\top\mathbf{M}\bm{\pi}-\mathbf{b}^\top\bm{\pi}.
    \label{costpi}
\end{equation}
To calculate the matching cost parallelly and reduce the computation in each independent connected component, we partition the quadratic affinity matrix $\mathbf{M}$ and vertex affinity matrix $\mathbf{B}$. We denote $\bm{\pi}^\top=[\bm{\pi}_c^\top,\bm{\pi}_m^\top]$, where $\bm{\pi}_c$ is in current connected component $\mathcal{G}_c$ and $\bm{\pi}_m$ is between the complement of the vertex sets, i.e., $\mathcal{G}_m=(\mathcal{V}_D\backslash \mathcal{V}^D_c,\mathcal{V}_T\backslash \mathcal{V}^T_c,\mathcal{E}^{DT}_m)$. Then, the matching cost is
\begin{equation}
\begin{split}
    \mathcal{L}(\bm{\pi}_c)=&-\bm{\pi}^\top\mathbf{M}\bm{\pi}-\mathbf{b}^\top\bm{\pi}\\
    =&-[\bm{\pi}_c^\top, {\bm{\pi}_m^{*\top}}]     
    \left[
    \begin{array}{cc}
        \mathbf{M}_c &\mathbf{M}_r\\
        \mathbf{M}_l &\mathbf{M}_m
    \end{array}
    \right]
    \begin{bmatrix}
        \bm{\pi}_c\\
        \bm{\pi}_m^*
    \end{bmatrix}\\
    &-[\mathbf{b}_c^\top, \mathbf{b}_m^\top]\begin{bmatrix}
        \bm{\pi}_c\\
        \bm{\pi}_m^*
    \end{bmatrix},\\
    =&-[\bm{\pi}_c^\top, {\bm{\pi}_m^{*\top}}]     
    \left[
    \begin{array}{c c}
        \mathbf{M}_c &\mathbf{M}_r\\
        \mathbf{M}_l &\mathbf{0}
    \end{array}
    \right]
    \begin{bmatrix}
        \bm{\pi}_c\\
        \bm{\pi}_m^*
    \end{bmatrix}-\mathbf{b}_c^\top\bm{\pi}_c,\\
    =&-\bm{\pi}_c^\top\mathbf{M}_c\bm{\pi}_c-2\bm{\pi}_m^{*\top}\mathbf{M}_l\bm{\pi}_c-\mathbf{b}_c^\top\bm{\pi}_c,\\
    =&-\bm{\pi}_c^\top\mathbf{M}_c\bm{\pi}_c+2\mathcal{L}_e(\bm{\pi}_c,{\bm{\pi}_m^{*}})-\mathbf{b}_c^\top\bm{\pi}_c,
    \label{costpic}
\end{split}
\end{equation}
where $\mathbf{M}_c\in \MBR^{|\mathcal{V}^D_c|\times|\mathcal{V}^T_c|}, \mathbf{b}_c\in\MBR^{|\mathcal{V}^D_c||\mathcal{V}^T_c|}$. Here, the matching cost contains the pairwise cost $\mathcal{L}_e(\bm{\pi}_c,{\bm{\pi}_m^{*}})$ that depends on the optimal solution $\bm{\pi}_m^{*}$, not available from connected component $\mathcal{G}_c$. To make it independent of the other components, we only consider the optimal solution $\widetilde{\bm{\pi}}_m^*$ given $\bm{\pi}_c$ instead the global optimal solution $\bm{\pi}_m^*$, i.e.,
\begin{equation}
    \begin{split}
\mathcal{L}_e(\bm{\pi}_c,{\bm{\pi}_m^{*}})\approx&\mathcal{L}_e({\widetilde{\bm{\pi}}_m^*|\bm{\pi}_c})=-\max_{\pi_m} \bm{\pi}_m^{\top}\mathbf{M}_l\bm{\pi}_c.
\end{split}
\end{equation}
Intuitively, it is to find the best matching between the edges in the detection graph and tracklet graph with the start vertices fixed to an existing set of matches $\bm{\pi}_c$. As shown in Fig.~\ref{fig:edgematch}, for each matched pair $(c_i^D,c_j^T)$ in $\bm{\pi}_c$, we adopt bipartite matching between edge set $\{e_{i,i'}\}$ and $\{e_{j,j'}\}$ that start from $c_i^D$ and $c_j^T$ respectively.

\begin{algorithm}[ht!]
\caption{Gated Search Tree (GST)}
    \label{alg:gst}
  
    \KwIn{$\mathbf{M}, \mathbf{B},\mathtt{iou}, \mathtt{KF},\sigma,\kappa$}
    \KwOut{$\mathbf{\Pi}$}
    {\small //\ Construct\ graph}\\

    \For{$(i.j) \in \mathtt{range}(n_d,n_t)$}{
    $\mathbf{A}_{i.j}\leftarrow\mathbb{I}\{{\mathtt{iou}_{i,j}>0 \land \mathtt{KF}_{i,j}>\kappa \land \mathbf{B}_{i,j}>\sigma}\}$}

    {\small //\ Find Independent Connected Components (Alg.~\ref{alg:FCS})}\\
    $\{\mathcal{G}_k\}\leftarrow\mathtt{FICC(\mathcal{G}(\mathbf{A}))}$\\
    {\small //\ Find Best Matching}\\
    \For{$\mathcal{G}_c \in \{\mathcal{G}_k\}$}{
    $\mathbf{M}_c, \mathbf{b}_c = \mathbf{M}[\{c\},\{c\}], \mathtt{vec}(\mathbf{B}[\{c\},\{c\}])$\\
    $\mathcal{L}(\bm{\pi}_c)=-\bm{\pi}_c^\top\mathbf{M}_c\bm{\pi}_c+2\mathcal{L}_e({\widetilde{\bm{\pi}}_m^*|\bm{\pi}_c})-\mathbf{b}_c^\top\bm{\pi}_c,$\\
    $\bm{\pi}_c^*\leftarrow\arg\min_{\bm{\pi}_c}\mathcal{L}(\bm{\pi}_c)$
    }
    $\mathbf{\Pi}\leftarrow \bigcup_{c=1}^{k} \bm{\pi}_c^*$\\
    \Return{$\mathbf{\Pi}$}

\end{algorithm}
\begin{algorithm}[]
    \caption{Find Independent Connected Components (FICC)}
        \label{alg:FCS}
        \SetKwProg{Fn}{def}{:}{end}
        \KwIn{$\mathcal{G}=(\mathcal{V}_D,\mathcal{V}_T,\mathcal{E}_{DT})$}
        \KwOut{$\{\mathcal{G}_c\}$}
        $c\leftarrow 0$; $\{\mathcal{G}_c\}\leftarrow\varnothing$\\
        \For{$v_p\in \mathcal{V}_D\cup\mathcal{V}_T$}{
            $\mathtt{visited}[v_p]\leftarrow\mathtt{False}$
        }

        \For{$v_p \in \mathcal{V}_D\cup\mathcal{V}_T$}{
            \If{$\neg\mathtt{visited}[v_p]$}{
                $\mathcal{E}_c^{DT},\mathcal{V}_c^D,\mathcal{V}_c^T \leftarrow \varnothing$\\
                \If {$v_p\in \mathcal{V}_D$}{
                $\mathcal{V}_c^D \leftarrow \mathcal{V}_c^D \cup\{v_p\}$}
                \ElseIf{$v_p\in \mathcal{V}_T$}{$\mathcal{V}_c^T \leftarrow \mathcal{V}_c^T \cup\{v_p\}$}
                $\mathtt{visit}(v_p)$\\
                $c \leftarrow c+1$}
        }
        \Fn{$\mathtt{visit}(v_p)$}{
            $\mathtt{visited}[v_p]\leftarrow\mathtt{True}$\\
            \For{$e_{p,q}\in \mathcal{E}_{DT}$}{
                $\mathcal{E}_c^{DT} \leftarrow \mathcal{E}_c^{DT}\cup\{e_{p,q}\}$\\
                \If{$\neg\mathtt{visited}[v_q]$}{
                    $\mathtt{visit}(v_q)$\\
                    \If {$v_q\in \mathcal{V}_D$}{
                $\mathcal{V}_c^D \leftarrow \mathcal{V}_c^D \cup\{v_q\}$}
                \ElseIf{$v_i\in \mathcal{V}_T$}{$\mathcal{V}_c^T \leftarrow \mathcal{V}_c^T \cup\{v_q\}$}
                }
            }

        }
        \Return{$\{\mathbf{G_c}\}$}
        
\end{algorithm}

Then, we discuss the time cost of the original QP solver and the GST algorithm:
\begin{proposition}[Original complexity]
    The quadratic programming Eq.~\ref{finalQP} can be solved in $O(n_d^3n_t^3)$ arithmetic operations.
\end{proposition}
\begin{proposition}
    The running time of Algorithm~\ref{alg:gst} in parallel mode is
    \begin{equation}
        T = c\cdot n_m|\mathcal{V}^T_1||\mathcal{V}^D_1|+\epsilon,
\end{equation}
where $c$ is a constant factor, $\epsilon$ represents low-order terms of $n$ and communication overhead between threads, $n_m=\max\{n_d,n_t\}$.
\end{proposition}
\begin{figure*}[!h] 
    \centering
    \vspace{-10pt}
    \includegraphics[width=\linewidth]{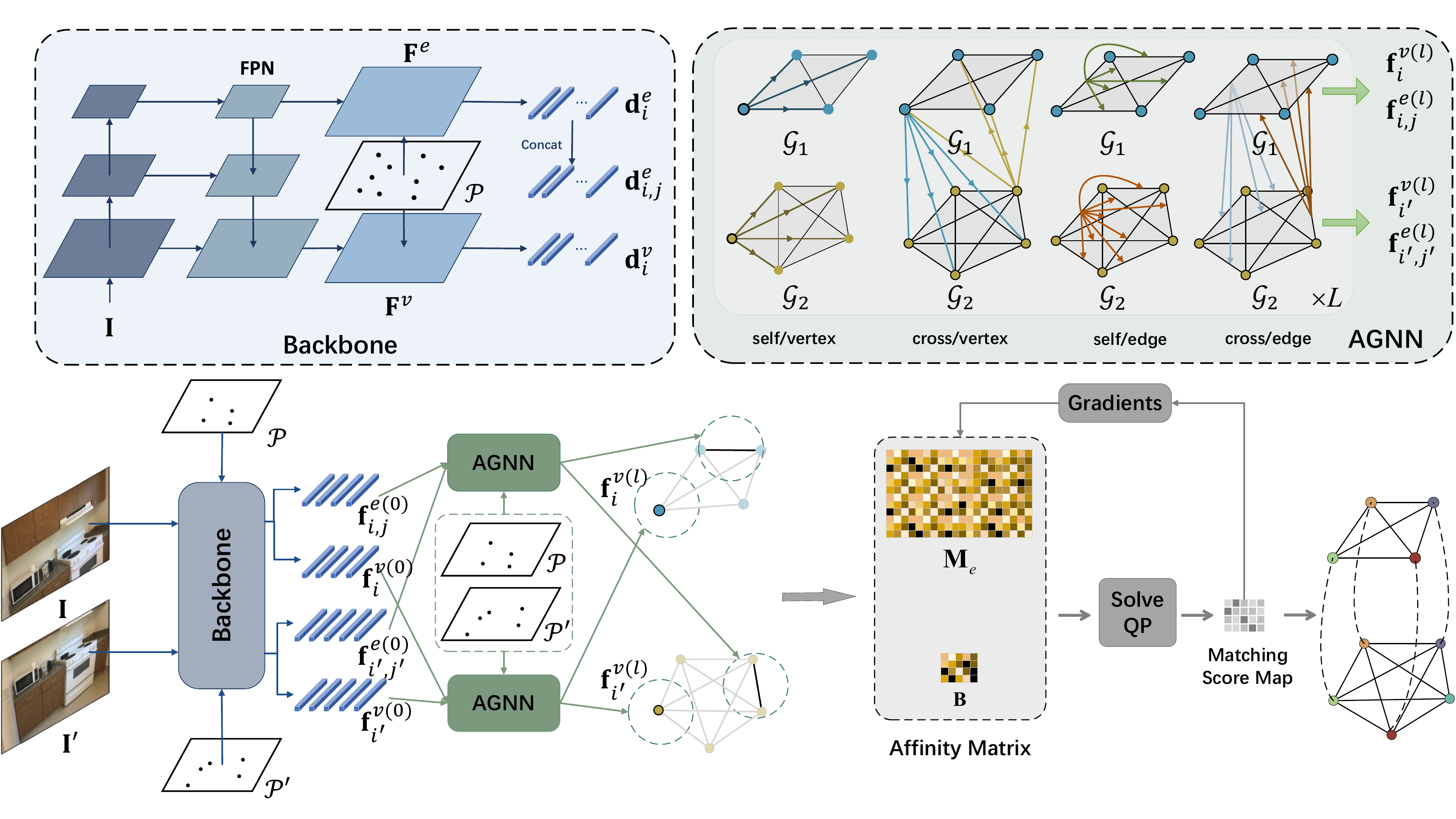}
    \vspace{-10pt}
    \caption{Pipeline of our image matching network, GMatcher. The backbone is an FPN-like module. The edge and vertex features are from the stride-8 and stride-2 feature maps respectively. Edge and vertex AGNN are operated independently. The learnable graph matching layer replaces the Sinkhorn layer in SuperGlue.}
    \vspace{-10pt}
    \label{fig:gmatcher}
\end{figure*}
\section{Learnable Graph Matching for Image Matching Task}
Besides the MOT task, our learnable graph matching method can be easily adapted to other data association tasks with slight modifications. In this section, we take the image matching task as an example. We formulate the image matching task as a graph matching problem between the keypoints in two images and utilize our learnable graph matching algorithm to build an end-to-end keypoint-based neural network.
\subsection{Problem Formulation}
Given keypoints $\mathcal{P}=\{\mathbf{p}_1,\mathbf{p}_2,\cdots,\mathbf{p}_m\}$, $\mathcal{P}'=\{\mathbf{p}'_1,\mathbf{p}'_2,\cdots,\mathbf{p}'_n\}$ on the image $I$ and $I'$ of the same scene respectively, where $\mathbf{p}_i=(x_i,y_i)$ is the keypoint position in image coordinates, the image matching task is to find the best matching between $\mathcal{P}$ and $\mathcal{P}'$ and thus estimate the relative camera pose $\mathbf{T}\in SE(3)$. 
The keypoints are often on the corners, textured areas, or the boundary of the objects, where the local features are relatively robust and less affected by illumination and viewing angle.
They can be derived from traditional methods, like SIFT~\cite{lowe2004distinctive}, or deep learning-based methods, like SuperPoint~\cite{detone2018superpoint}. And the descriptor $\mathbf{d}_i \in \mathbb{R}^c$ is a $c$-dimentional local discriminative feature, corresponding to the keypoint $\mathbf{p}_i$. In this paper, we extract the features end-to-end, with only the positions of the keypoints from the off-the-shelf neural network. \par
In our end-to-end graph matching neural network, called \emph{GMatcher}, we take two images and the keypoints on each image as the input, solving the graph matching problem (Eq.~\ref{finalQP}) from $\mathcal{P}$ and $\mathcal{P}'$, and we finally obtain the assignment matrix $\mathbf{\Pi} \in \mathbb{R}^{m\times n}$ to represent the matching between two keypoint sets. 
\subsection{End-to-end Graph Matching Network for Image Matching}
Our method is mainly based on SuperGlue~\cite{sarlin2020superglue}, which utilizes the attentional graph neural network (AGNN) module to aggregate long-range dependencies. However, compared with graph matching, although SuperGlue stacks self- and cross-attention modules many times to fuse the intra- and inter-image information, it does not explicitly define the intra-image graph and consider the edge similarities between images in keypoint matching. \par
To better utilize the high-order information, i.e., edge in the intra-image graph, we use FPN-like~\cite{lin2017feature} backbone to extract multiscale features. The stride-2 feature map and stride-8 feature map are used to extract vertex feature $\mathbf{d}_i^v$ and edge feature $\mathbf{d}_{i,j}^e$ respectively. Note that the edge feature $\mathbf{d}_{i,j}^e$ is the concatenation of the endpoints' feature $\mathbf{d}_i^e$ and $\mathbf{d}_j^e$ on stride-8 feature map. The feature $\mathbf{d}_i$ on each keypoint $\mathbf{p}_i$ is extracted from the feature map that is restored to the original image resolution using bilinear interpolation.\par
Like the position embedding in the transformer, we use MLP to encode the position information into the vertex feature $\mathbf{d}_i^v$ and edge feature $\mathbf{d}_{i,j}^e$, i.e.,
\begin{equation}
    \begin{aligned}
    \label{eq:keypoint-encoder}
    \mathbf{f}_i^v &= \mathbf{d}_i^v + \text{MLP}_{\text{pos}}(\mathbf{p}_i),\\
    \mathbf{f}_{i,j}^e &= \mathbf{d}_{i,j}^e + [\text{MLP}_{\text{pos}}(\mathbf{p}_i),\text{MLP}_{\text{pos}}(\mathbf{p}_j)],
    \end{aligned}
\end{equation}
where, $[\cdot,\cdot]$ denotes concatenation.\par
Then, similar to the AGNN module in SuperGlue, we conduct self-attentional and cross-attentional message passing for $l$ times in vertex features and edge features separately. The detailed design can be referred to SuperGlue~\cite{sarlin2020superglue}. We use the output vertex features and edge features of AGNN to construct the final graphs $\mathcal{G}_1=(\mathcal{V}=\{\mathbf{f}_i^{v(l)}\},\mathcal{E}=\{\mathbf{f}_{i,j}^{e(l)}\}))$ and $\mathcal{G}_2=(\mathcal{V}=\{\mathbf{f}_{i'}^{v(l)}\},\mathcal{E}=\{\mathbf{f}_{i',j'}^{e(l)}\}))$ for two images and matching with our differentiable graph matching layer mentioned in Sec.~\ref{sec:diffgm}.
   
\section{Experiments on MOT task}

\subsection{Datasets and Evaluation Metrics}
We carry out the experiments on MOT16~\cite{milan2016mot16} and MOT17~\cite{milan2016mot16} benchmark. The videos in this benchmark were taken under various scenes, light conditions and frame rates. Occlusion, motion blur, camera motion and distant pedestrians are also crucial problems in this benchmark. Among all the evaluation metrics, Multiple Object Tracking Accuracy (MOTA)~\cite{kasturi2008framework} and ID F1 Score (IDF1)~\cite{ristanieccvw2016} are the most general metrics in the MOT task. Since MOTA is mostly dominated by the detection metrics false positive and false negative, and our graphing matching method mainly tries to tackle the associations between detected objects, we pay more attention to IDF1 than the MOTA metric. Moreover, a newly proposed metric Higher Order Tracking Accuracy (HOTA)~\cite{luiten21hota}, emphasizing the balance of object detection and association, becomes one of the official metrics of MOT benchmarks.\\
\begin{table*}[!t]
    \centering

    \tabcolsep=0.12cm

        \begin{tabular}{l |c|c c c c c c c c c c c}
         \toprule
         Methods &Refined Det& IDF1 $\uparrow$ &HOTA$\uparrow$& MOTA $\uparrow$ &MT$\uparrow$ &ML$\downarrow$ & FP $\downarrow$ & FN $\downarrow$ & IDS $\downarrow$  &AssA$\uparrow$ &DetA$\uparrow$ &LocA$\uparrow$\\ [0.5ex] 
         \midrule
         \multicolumn{13}{c}{MOT17} \\
         \midrule
         Tracktor++ (O)  \cite{bergmann2019tracking}&Tracktor & 52.3&42.1& 53.5&19.5&36.6 & 12201 & 248047 & 2072 &41.7 &42.9 &80.9\\
         Tracktor++v2 (O)  \cite{bergmann2019tracking}&Tracktor& 55.1&44.8 & 56.3&21.1&35.3  & 8866 & 235449 & 1987&45.1&44.9 &81.8\\
         GNNMatch (O) \cite{papakis2020gcnnmatch}&Tracktor  &56.1 &45.4&57.0&23.3&34.6 &12283 	&228242 &1957&45.2 	&45.9  &81.5\\
         GSM\_Tracktor (O) \cite{liugsm}&Tracktor &57.8 &45.7&56.4&22.2&34.5&14379 &230174 &1485 &47.0 &44.9 &80.9\\
         CTTrackPub (O) \cite{zhou2020tracking} &CenterTrack &59.6&48.2&61.5  &\textbf{26.4}  &\textbf{31.9} &14076 &200672 &2583 &47.8 &49.0 &81.7\\
         BLSTM-MTP-T (O)~\cite{Kim_2021_CVPR} &Tracktor&60.5&-&55.9&20.5&36.7&\textbf{8663}&238863&\textbf{1188}&-&-&-\\
         TADAM (O)~\cite{Guo_2021_CVPR} &Tracktor&58.7&-&59.7&-&-&9676 &216029&1930&-&-&-\\
         ArTIST-C (O)~\cite{Saleh_2021_CVPR} &CenterTrack&59.7&48.9&\textbf{62.3}&29.1&34.0&19611&191207&2062&48.3&\textbf{50.0}&81.4\\
         \bf{GMTracker(Ours)} (O) &Tracktor& 63.8 &49.1& 56.2  &21.0 &35.5 &8719 & 236541 & 1778 &53.9 &44.9 &81.8 \\
         \bf{GMT\_CT(Ours)} (O) &CenterTrack& \textbf{66.9}&\textbf{52.0}& 61.5 &26.3 &32.1 & 14059 & \textbf{200655} & 2415 &\textbf{55.1} &49.4 &\textbf{81.8} \\
         \midrule
         MPNTrack~\cite{braso2020learning} &Tracktor& 61.7 &49.0& 58.8&28.8&33.5 & 17413 &213594 &1185 &51.1&47.3&81.5\\
         Lif\_TsimInt~\cite{hornakova2020lifted}&Tracktor&65.2 &50.7&58.2 &28.6&33.6 &16850 &217944 &\textbf{1022} &54.9&47.1&81.5\\
         LifT~\cite{hornakova2020lifted}&Tracktor &65.6 &51.3&60.5 &27.0&33.6&14966 &206619 &1189&54.7&48.3&81.3\\
         LPC\_MOT~\cite{Dai_2021_CVPR} &Tracktor&66.8&51.5&59.0&29.9&33.9&23102&206948&1122&56.0	&47.7	&80.9\\
         ApLift~\cite{hornakova2021making} &Tracktor&65.6&51.1&60.5&\textbf{33.9}&\textbf{30.9}&30609&190670&1709&53.5	&49.1	&80.7\\
         \bf{GMT\_simInt (Ours)} &Tracktor&  65.9&51.1& 59.0 &29.0 &33.6&  20395 & 209553 & 1105 &55.1 &47.6 &81.2 \\
         \bf{GMT\_VIVE (Ours)} &Tracktor&65.9 &51.2 &60.2 &26.5 &33.2 &\textbf{13142} &209812 &1675 &55.1 &47.8 &81.3\\
         \bf{GMTCT\_simInt (Ours)}&CenterTrack &\textbf{68.7} &\textbf{54.0}&\textbf{65.0} &29.4 &31.6&18213 &\textbf{177058} &2200 &\textbf{56.4} &\textbf{52.0} &\textbf{81.5}\\
         \midrule
         \multicolumn{13}{c}{MOT16} \\
         \midrule
         Tracktor++v2 (O)  \cite{bergmann2019tracking} &Tracktor&54.9 &44.6&56.2&20.7&35.8 &2394 	&76844 	&617 &44.6&44.8&82.0\\
         GNNMatch (O) \cite{papakis2020gcnnmatch} &Tracktor&55.9  &44.6&56.9&22.3&35.3&3235 	&74784 	&564 &43.7&45.8&81.7\\
         GSM\_Tracktor (O)\cite{liugsm} &Tracktor &58.2&45.9 &57.0  &22.0&34.5 &4332 &73573 &\textbf{475} &46.7&45.4&81.1\\
         TADAM (O)~\cite{Guo_2021_CVPR} &Tracktor&59.1 &-&59.5 &-&-&2540 &71542 &529&-&-&-\\
         ArTIST-C (O)~\cite{Saleh_2021_CVPR} &CenterTrack&61.9&49.8&\textbf{63.0}&29.1&33.2&7,420&59,376&635&49.5&\textbf{50.6}&81.0\\
         \bf{GMTracker(Ours)} (O)  &Tracktor&  63.9&48.9& 55.9 &20.3 &36.6 & \textbf{2371}& 77545 & 531  &53.7 &44.6 &\textbf{82.1}\\
         \bf{GMT\_CT (Ours)} (O)&CenterTrack &  \textbf{68.6} &\textbf{53.1}&62.6 &\textbf{26.7} &\textbf{31.0} &  5104 &  \textbf{62377} & 787 &\textbf{56.3}&50.4  &81.8 \\
         \midrule
         MPNTrack~\cite{braso2020learning}&Tracktor  & 61.7&48.9& 58.6  &27.3&34.0 & 4949 &70252 &354&51.1&47.1&81.7\\
         Lif\_TsimInt~\cite{hornakova2020lifted} &Tracktor &64.1 &49.6&57.5 &25.4&34.7&\textbf{4249} &72868 &\textbf{335}&53.3&46.5&\textbf{81.9}\\
         LifT~\cite{hornakova2020lifted}&Tracktor&64.7 &50.8&61.3 &27.0&34.0 &4844 &65401 &389&53.1&48.9&81.4\\
         LPC\_MOT~\cite{Dai_2021_CVPR} &Tracktor&67.6&51.7&58.8&27.3&35.0&6167&68432&435&56.4&47.6&81.3\\
         ApLift~\cite{hornakova2021making} &Tracktor&66.1&51.3&61.7&\textbf{34.3}&31.2&9168&60180&495&53.2&49.8&80.7	\\
         \bf{GMT\_simInt (Ours)} &Tracktor&  66.2 &51.2& 59.1  &27.5 &34.4 &  6021 &  68226 & 341 &55.1 &47.7 &81.5 \\
         \bf{GMT\_VIVE (Ours)}&Tracktor&66.6 &51.6&61.1 &26.7 &33.3&3891 &66550 &503 &55.3 &48.5 &81.5\\
         \bf{GMTCT\_simInt (Ours)}&CenterTrack  &\textbf{70.6}&\textbf{55.2} &\textbf{66.2}&29.6 &\textbf{30.4} &6355 &\textbf{54560} &701 &\textbf{57.8} &\textbf{53.1} &81.5\\
         \bottomrule
        \end{tabular}
    
    \caption{Comparison with state-of-the-art methods under `public detection' setting on MOT16 and MOT17 \emph{test} set. (O) denotes online methods. (O$^*$) denotes near-online methods.}
    \label{tab:mot}
    \end{table*}
\begin{table}[h]
    \begin{center}
    \resizebox{\columnwidth}{!}{
\begin{tabular}{ l | c c c c c c c}
\toprule
Tracker & MOTA$\uparrow$ & IDF1$\uparrow$ & HOTA$\uparrow$ & FP$\downarrow$ & FN$\downarrow$ & IDs$\downarrow$ & FPS$\uparrow$\\
\midrule
DAN \cite{sun2019deep} & 52.4 & 49.5 & 39.3 & 25423 & 234592 & 8431 & \textless 3.9\\ 
Tube\_TK \cite{pang2020tubetk} & 63.0 & 58.6 & 48.0 & 27060 & 177483 & 4137 & 3.0\\
MOTR \cite{zeng2021motr} & 65.1 & 66.4 & - & 45486 & 149307 & 2049 & -\\
CTracker \cite{peng2020chained} & 66.6 & 57.4 & 49.0 & 22284 & 160491 & 5529 & 6.8\\
CenterTrack \cite{zhou2020tracking} & 67.8 & 64.7 & 52.2 & \textbf{18498} & 160332 & 3039 & 17.5\\
QuasiDense \cite{pang2021quasi} & 68.7 & 66.3 & 53.9 & 26589 & 146643 & 3378 & 20.3\\
TraDes \cite{wu2021track} & 69.1 & 63.9 & 52.7 & 20892 & 150060 & 3555 & 17.5\\
MAT \cite{han2020mat} & 69.5 & 63.1 & 53.8 & 30660 & 138741 & 2844 & 9.0\\
SOTMOT \cite{zheng2021improving} & 71.0 & 71.9 & - & 39537 & 118983 & 5184 & 16.0\\
TransCenter \cite{xu2021transcenter} & 73.2 & 62.2 & 54.5 & 23112 & 123738 & 4614 & 1.0\\
GSDT \cite{wang2020joint} & 73.2 & 66.5 & 55.2 & 26397 & 120666 & 3891 & 4.9\\
Semi-TCL \cite{li2021semi} & 73.3 & 73.2 & 59.8 & 22944 & 124980 & 2790 & -\\
FairMOT \cite{zhang2020fairmot} & 73.7 & 72.3 & 59.3 & 27507 & 117477 & 3303 & 25.9\\
RelationTrack \cite{yu2021relationtrack} & 73.8 & 74.7 & 61.0 & 27999 & 118623 & 1374 & 8.5\\
PermaTrackPr \cite{tokmakov2021learning} & 73.8 & 68.9 & 55.5 & 28998 & 115104 & 3699 & 11.9\\
CSTrack \cite{liang2020rethinking} & 74.9 & 72.6 & 59.3 & 23847 & 114303 & 3567 & 15.8\\
TransTrack \cite{sun2020transtrack} & 75.2 & 63.5 & 54.1 & 50157 & 86442 & 3603 & 10.0\\
FUFET \cite{shan2020tracklets} & 76.2 & 68.0 & 57.9 & 32796 & 98475 & 3237 & 6.8\\
SiamMOT \cite{liang2021one} & 76.3 & 72.3 & - & - & - & - & 12.8\\
CorrTracker \cite{wang2021multiple} & 76.5 & 73.6 & 60.7 & 29808 & 99510 & 3369 & 15.6\\
TransMOT \cite{chu2021transmot} & 76.7 & 75.1 & 61.7 & 36231 & 93150 & 2346 & 9.6\\
ReMOT \cite{yang2021remot} & 77.0 & 72.0 & 59.7 & 33204 & 93612 & 2853 & 1.8\\
ByteTrack \cite{zhang2022bytetrack}& 80.3 & 77.3 & 63.1 & 25491 & \textbf{83721} & 2196 &\textbf{29.6}\\
\textbf{GMTracker (Ours)}&\textbf{80.6} &\textbf{79.8} &\textbf{64.9} &22119 &86061  &\textbf{1197} &13.2\\
\bottomrule
\end{tabular}}
      \end{center}
    \caption{Comparisons with SOTA methods under 'private detection' setting on MOT17~\cite{milan2016mot16} test set.}
  \label{tab:mot17pri}
  \end{table}

\begin{table}[h]
    \begin{center}
       \resizebox{\columnwidth}{!}{
\begin{tabular}{ l | c c c c c c c}

\toprule
Tracker & MOTA$\uparrow$ & IDF1$\uparrow$ & HOTA$\uparrow$ & FP$\downarrow$ & FN$\downarrow$ & IDs$\downarrow$ & FPS$\uparrow$\\
\midrule
MLT \cite{zhang2020multiplex} & 48.9 & 54.6 & 43.2 & 45660 & 216803 & 2187 & 3.7\\
FairMOT \cite{zhang2020fairmot} & 61.8 & 67.3 & 54.6 & 103440 & 88901 & 5243 & 13.2\\
TransCenter \cite{xu2021transcenter} & 61.9 & 50.4 & - & 45895 & 146347 & 4653 & 1.0\\
TransTrack \cite{sun2020transtrack} & 65.0 & 59.4 & 48.5 & 27197 & 150197 & 3608 & 7.2\\
CorrTracker \cite{wang2021multiple} & 65.2 & 69.1 & - & 79429 & 95855 & 5183 & 8.5\\
Semi-TCL \cite{li2021semi} & 65.2 & 70.1 & 55.3 & 61209 & 114709 & 4139 & -\\
CSTrack \cite{liang2020rethinking} & 66.6 & 68.6 & 54.0 & 25404 & 144358 & 3196 & 4.5\\
GSDT \cite{wang2020joint} & 67.1 & 67.5 & 53.6 & 31913 & 135409 & 3131 & 0.9\\
SiamMOT \cite{liang2021one} & 67.1 & 69.1 & - & - & - & - & 4.3\\
RelationTrack \cite{yu2021relationtrack} & 67.2 & 70.5 & 56.5 & 61134 & 104597 & 4243 & 2.7\\
SOTMOT \cite{zheng2021improving} & 68.6 & 71.4 & - & 57064 & 101154 & 4209 & 8.5\\
ByteTrack \cite{zhang2022bytetrack} & 77.8 & 75.2 & 61.3 & 26249 & \textbf{87594} & \textbf{1223} & \textbf{17.5}\\
\textbf{GMTracker (Ours)} & \textbf{77.8} & \textbf{76.7} & \textbf{62.9} & \textbf{24409} &88911 & 1331 & 5.6\\
\bottomrule
\end{tabular}}
      \end{center}
    \caption{Comparisons with SOTA methods under 'private detection' setting on MOT20~\cite{dendorfer2020mot20} test set.}
  \label{tab:mot20pri}
  \end{table}

\begin{table*}
    \begin{center}
      \begin{tabular}{ccccc|cc|ccccc}
        \toprule
        GM &App. Enc. &GCN &Geo &Inter. & IDF1 $\uparrow$& MOTA $\uparrow$& MT $\uparrow$& ML $\downarrow$& FP $\downarrow$& FN $\downarrow$& ID Sw. $\downarrow$\\
        \midrule
        &&& & &  68.1 &  62.1  &  556 &  371  &  1923  &  124480  &  1135  \\
        \checkmark&&& &   &  70.0  &  62.3 & 555 &  374  &  1735  & 124292 &1128\\
        \checkmark&&&\checkmark &   &70.2 &  62.2  &  555 & 374 & 1744  & 124301  &  1140 \\
        \checkmark&\checkmark&& & &70.4 & 62.3  &  554 &  375  &  1741  & 124298  & 1058  \\
        \checkmark&\checkmark&\checkmark& &  &70.6 &  62.2  &  556 &  374  &  1748  &  124305  &  1399  \\
        \checkmark&\checkmark&\checkmark&\checkmark & &71.5 & 62.3  &  555 &  375  & 1741  &  124298  & 1017  \\
        \midrule
        &&&  &\checkmark &  68.9 &  62.9  & 678 &  361  &  11440  &  112853  &  723  \\
        \checkmark&&&   &\checkmark  &71.6 &  64.0  &  669 &  365  &  7095  &  113392  & 659  \\
        \checkmark&&&\checkmark& \checkmark   &71.7 & 64.0  & 666 &  364  & 6816  & 113778  & 724 \\
        \checkmark&\checkmark&&  &\checkmark  &72.0 &  64.2  & 671 &  368  &  7701  &  112370  &  627  \\
        \checkmark&\checkmark&\checkmark&    &\checkmark  &72.1 &  63.3  &  676 &  364  &  10888  &  111869  & 716  \\
        \checkmark&\checkmark&\checkmark&\checkmark &\checkmark  &73.0 &  63.8  & 672 &  361  &  9579  & 111683  &  570  \\
        \bottomrule
      \end{tabular}
      \end{center}
    \caption{Ablation studies on different proposed components on MOT17 \emph{val} set.}
  \label{tbl-ablation}
  \end{table*}
\subsection{Implementation Details}
\noindent\textbf{Training.}
Following other MOT methods \cite{braso2020learning,hornakova2020lifted}, we adopt Tracktor~\cite{bergmann2019tracking} to refine the public detections. We use a ResNet50 \cite{he2016deep} backbone followed by a global average pooling layer and a fully connected layer with 512 channels, as the ReID network used for feature extraction. The output ReID features are further normalized with the $l_2$ normalization. We pre-train the ReID network on Market1501~\cite{market_dataset}, DukeMTMC~\cite{ristanieccvw2016} and CUHK03~\cite{cuhk03_dataset} datasets jointly. The parameters of the ReID network will be frozen after pre-training. Then we add two trainable fully connected layers with 512 channels to get appearance features. All the ReID network training settings follow MPNTrack~\cite{braso2020learning}.
Our implementation is based on PyTorch~\cite{paszke2019pytorch} framework. We train our model on an NVIDIA RTX 2080Ti GPU. Adam~\cite{kingma2014adam} optimizer is applied with $\beta_1=0.9$ and $\beta_2=0.999$. The learning rate is  5$\times$10$^{-5}$ and weight decay is 10$^{-5}$. The temperature $\tau$ in Eq. \ref{eq:softmax} is 10$^{-3}$. 

\noindent\textbf{Inference.}
Our inference pipeline mostly follows DeepSORT \cite{wojke2017simple}, except that we use general graphing matching instead of bipartite matching for the association. As in DeepSORT, we set the motion gate $\kappa$ as 9.4877, which is at the 0.95 confidence of the inverse $\chi^2$ distribution. The feature similarity threshold $\sigma$ is set to 0.6 in the videos taken by the moving camera, and 0.7 when we use geometric information in the cross-graph GCN module for videos taken by the static camera. The \emph{max age} $\delta$ is 100 frames.
\par
\noindent\textbf{Post-processing.} To compare with other state-of-the-art offline methods, we perform a linear interpolation within the tracklet as post-processing to compensate for the missing detections, following \cite{braso2020learning,hornakova2020lifted}. This effectively reduces the false negatives introduced by upstream object detection algorithm.\par
\subsection{Comparison with State-of-the-Art Methods}
We compare our GMTracker with other state-of-the-art methods on MOT16 and MOT17 test sets. As shown in Table \ref{tab:mot}, when we apply Tracktor~\cite{bergmann2019tracking} to refine the public detection, the \emph{online} \gm\  achieves 63.8 IDF1 on MOT17 and 63.9 IDF1 on MOT16, outperforming the other online trackers. To compare with CenterTrack~\cite{zhou2020tracking}, we use the same detections, called GMT\_CT, and the IDF1 is 66.9 on MOT17 and 68.6 on MOT16.\par

With the simple linear interpolation, called GMT\_simInt in Table \ref{tab:mot}, we also outperform the other \emph{offline} state-of-the-art trackers on IDF1. With exactly the same visual inter- and extrapolation as LifT~\cite{hornakova2020lifted}, called GMT\_VIVE in Table~\ref{tab:mot},\ the MOTA is comparable with LifT. After utilizing the CenterTrack detections and linear interpolation, the GMTCT\_simInt improves the SOTA on both MOT16 and MOT17 datasets.\par
In Table~\ref{tab:mot}, we use the public detections, Faster R-CNN, DPM, and SDP. Many of new SOTA methods use their private detection results. To fairly compare with them, we also report results under the 'private detection' setting on MOT17 and MOT20 datasets. We use YOLOX~\cite{ge2021yolox} object detector. The results are shown in Table~\ref{tab:mot17pri} and Table~\ref{tab:mot20pri}. Compared with newly published SOTA methods until 2022, we have 2.5 and 1.5 IDF1 improvements on MOT17 and MOT20.\par

\subsection{Ablation Study}
\label{sec:ablation}

\begin{table}
    \begin{center}
      \begin{tabular}{cc|cc}
        \toprule
        Train w/ GM & Inference w/ GM & IDF1 & MOTA \\
        \midrule
        & &69.5 &62.1\\
        &\checkmark &70.2 &62.3\\
        \checkmark &\checkmark &71.5 &62.3\\
        \bottomrule
      \end{tabular}
      \end{center}
    \caption{Ablation study on the graph matching layer.}
  \label{tab-rmgmlayer}
  \end{table}

\begin{table}
    \center
      \begin{tabular}{l|cc}
        \toprule
        \multicolumn{1}{c|}{Methods} & IDF1 & MOTA  \\
        \midrule
        Last Frame  &  64.3 &  62.2  \\
        Moving Average $\alpha=0.2$ &  69.8&  62.4 \\
        Moving Average $\alpha=0.5$ &  70.0&  62.4  \\
        Moving Average $\alpha=0.8$ &  70.6&  62.4 \\
        Mean  &  71.5 &  62.3    \\
        \bottomrule
      \end{tabular}
    \caption{Ablation studies on different intra-tracklet feature aggregation methods.}
  \label{tbl:traagg}
  \end{table}
  \begin{table}
    \center
      \begin{tabular}{l|ccccc}
        \toprule
        Max age (frames) &30 &50& 80 &100 &150\\
        \midrule
        DeepSORT  & 67.2 &67.9  &68.1 &68.1 &67.3\\
        Graph Matching  &  69.2 & 70.5 &71.4 &71.5 &71.8\\
        \bottomrule
      \end{tabular}
    \caption{The influence of \emph{max age} $\delta$ on IDF1.}
  \label{tbl-ablation2}
  \end{table}
  \begin{table}[h]
    \center
      \begin{tabular}{l|ccccc}
        \toprule
         &IDF1 &MOTA& IDs\\
        \midrule
        vertex GCN\_1layer  &71.5 &62.3 &1017\\
        vertex GCN\_2layers &71.1 &62.3 &1055\\
        vertex GCN\_2layers\_res &71.8 &62.3 &1025\\
        vertex GCN\_3layers\_res &71.8 &62.3 &1031\\
        vertex+edge GCN\_2layers &72.0 &62.3 &1074\\
        vertex+edge GCN\_2layers\_res &71.7 &62.3 &1113\\
        \bottomrule
      \end{tabular}
    \caption{The ablation study about the GNN module. *\_res means adding a residual block for each GCN layer.}
  \label{tbl:gnn_abl}
  \end{table}
We conduct ablation studies of the proposed components in our method on the MOT17 dataset. Following \cite{braso2020learning}, we divide the training set into three parts for three-fold cross-validation, called MOT17 \emph{val} set, and we conduct the experiments under this setting both in the ablation study section and the discussions section. We ablate each component we propose: (i) graph matching module built as a QP layer (GM); (ii) MLP trained on MOT dataset to refine the appearance features (App. Enc.); (iii) the cross-graph GCN module (GCN) with and without using geometric information (Geo); (iv) the linear interpolation method between the same object by the time (Inter.).

As shown in Table \ref{tbl-ablation}, compared with the DeepSORT baseline (the first row), which associates the detections and the tracklets based on Hungarian Algorithm, our method without training gets a gain of 1.9 IDF1, and a gain of 2.7 IDF1 and 1.1 MOTA with the linear interpolation. The results show the effectiveness of the second-order information in the graph. 

Appearance feature refinement and GCN improve about 0.6 IDF1 compared to the untrained model. Geometric information provides about 1.0 additional gain on IDF1, which highlights the importance of geometric information in the MOT task. Finally, compared with the baseline, our method achieves about 3.4 and 0.2 improvements on IDF1 metric and MOTA metric, respectively. With interpolation, the gain becomes even larger: about 4.1 improvements on IDF1 and 0.9 on MOTA.

Table \ref{tab-rmgmlayer} shows the effectiveness of our differentiable graph matching layer the importance of training all components in our tracker jointly. We get the gain of 1.3 and 2.0 IDF1 compared with only removing the graph matching layer in training stage and in both training and inference stage, respectively. 
\begin{figure}
    \includegraphics[width=\columnwidth]{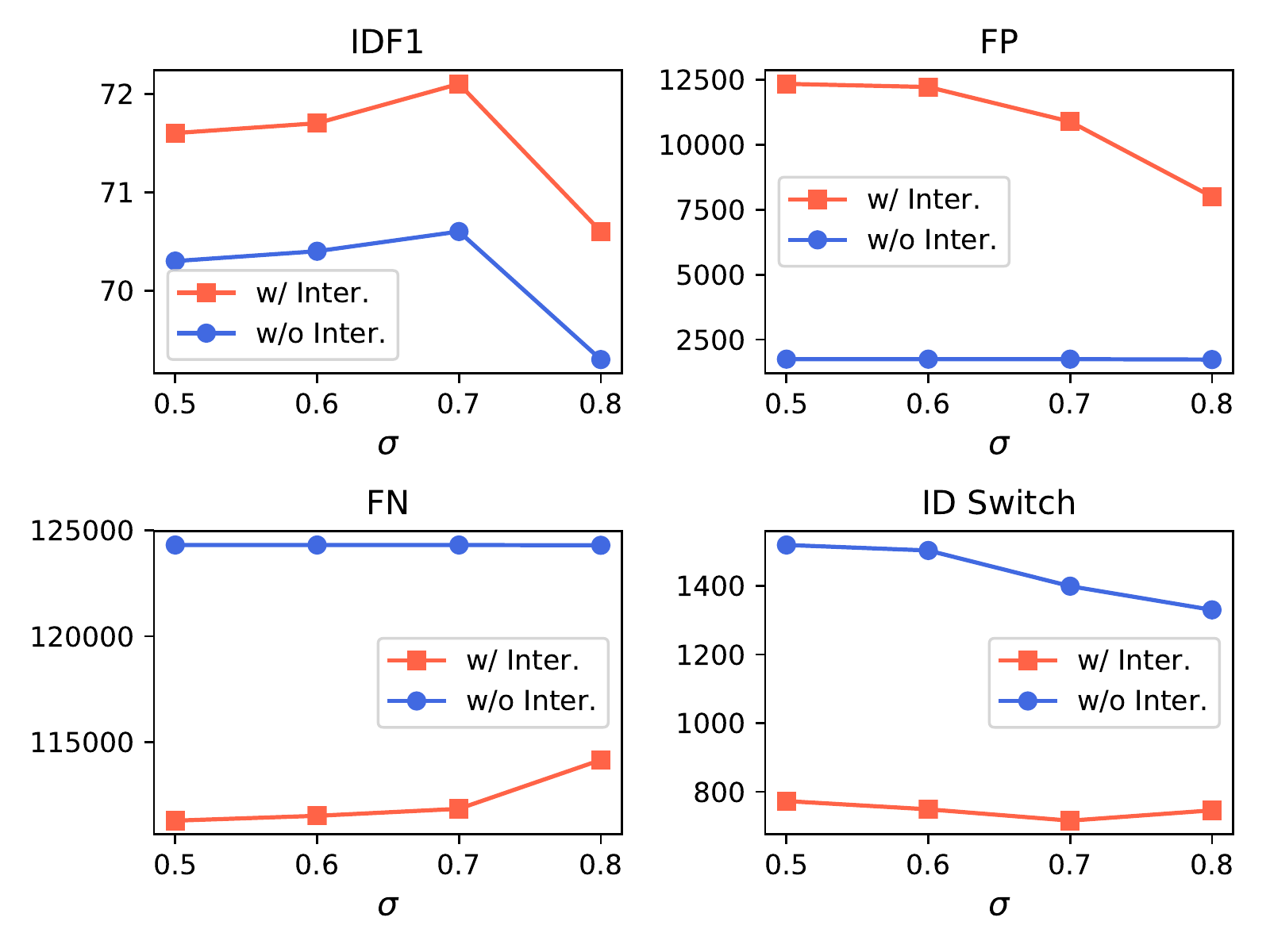}
    \caption{Results on IDF1, FP, FN and ID Switch metrics under different threshold $\sigma$ of the feature similarity to create a new tracklet.}
    \vspace{-0.2cm}
    \label{fig:thr}
    \vspace{-3pt}
\end{figure}
\begin{table}
    \begin{center}
        \resizebox{\columnwidth}{!}{
      \begin{tabular}{c|cc|ccccc}
        \toprule
         &IDF1 & MOTA & MT & ML & FP & FN & ID Sw.\\
        \midrule
        Baseline &68.1 &  62.1  &  556 &  371  &  1923  &  124480  &  1135  \\
        Ours &71.5 & 62.3  &  555 &  375  & 1741  &  124298  & 1017  \\
        Oracle &77.2 &62.6 &545 &368 &1730 &124287 &14 \\
        \bottomrule
      \end{tabular}}
      \end{center}
    \caption{Comparison between the baseline, our GMTracker and the Oracle tracker on MOT17 \emph{val} set.}
  \label{tbl-oracle}
  \end{table}

\subsection{Discussions}
In this part, we discuss two main design choices of our method on the MOT17 \emph{val} set. When we construct the tracklet graph, there are some different intra-tracklet feature aggregation methods. Moreover, how to create and delete a tracklet is important for an online tracker. Besides, the oracle experiment shows the upper bound performance of our learnable graph matching method.

\noindent{\bf Intra-tracklet feature aggregation.} In the tracklet graph $\MCG_T$, each vertex represents a tracklet. And the vertex feature $\mathbf{a}_T^{j}$ is the aggregation of the appearance features of all detections in tracklet $T_j$. Here, we compare several aggregation methods, including mean, moving average and only using the last frame of the tracklet. The results are shown in Table \ref{tbl:traagg}. The IDF1 is 7.2 lower when only using the last frame of the tracklet. The results also reveal that when we utilize all the frame information, no matter using the simple average or the moving average, their impact is not significant. To make our method simple and effective, we finally use the simple average method to aggregate the appearance features within a tracklet.\par
\noindent{\bf GNN layers.} We discuss the number of GNN layers and whether to use the edge GNN or not. As shown in Table~\ref{tbl:gnn_abl}. The results indicate that with deeper GCN, the results can be slightly improved. However, the residual connection between layers is important. The edge GCN module can only have 0.2 (72.0 vs. 71.8) IDF1 improvement and takes more computation cost.

\noindent{\bf Tracklet born and death strategies.} 
In most of the online tracking methods, one of the core strategies is how to create and delete a tracklet. In our GMTracker, we mostly follow DeepSORT, but we also make some improvements to make these strategies more suitable for our approach, as described in Section \ref{sec:tarcker}.
 Among the three criteria to create a new tracklet, we find that the threshold $\sigma$ is the most sensitive hyperparameter in our method. We conduct experiments with different $\sigma$, and its influence on IDF1, FP, FN and ID Switch is shown in Fig.~\ref{fig:thr}. 
As for removing a trajectory from association candidates, our basic strategy is that if the tracklet has not been associated with any detections in $\delta$ frames, the tracklet will be removed and not be matched anymore.\par
Table \ref{tbl-ablation2} shows in our method, that larger \emph{max age} $\delta$, which means more tracklet candidates, yields a better IDF1 score. It shows the effectiveness of our method from another aspect that our GMTracker can successfully match the tracklets that disappeared about five seconds ago. On the contrary, when the \emph{max age} increases to 150 frames, the IDF1 will drop 0.8 using DeepSORT, which indicates our method can deal with long-term tracklet associations better.\par 
\noindent{\bf Comparison with the Oracle Tracker.}
To explore the upper bound of the association method, we compare our method with the ground truth association, called the Oracle tracker. The results on MOT17 \emph{val} set are shown in Table \ref{tbl-oracle}. There is a gap of 5.7 IDF1 and about 1000 ID Switches between our online GMTracker and the Oracle tracker. Another observation is that on some metrics, which are extremely relevant to detection results, like MOTA, FP and FN, the gaps between the baseline, our method and the Oracle tracker are relatively small. That is why we mainly concern with the metrics reflecting the association results, such as IDF1 and ID Switch.
\begin{figure*}[!h]
    \centering
    \subfloat[Detection misses and resurfaces by occlusion.]{
    \begin{minipage}[]{0.5\linewidth}
    \centering
    \includegraphics[width=\linewidth]{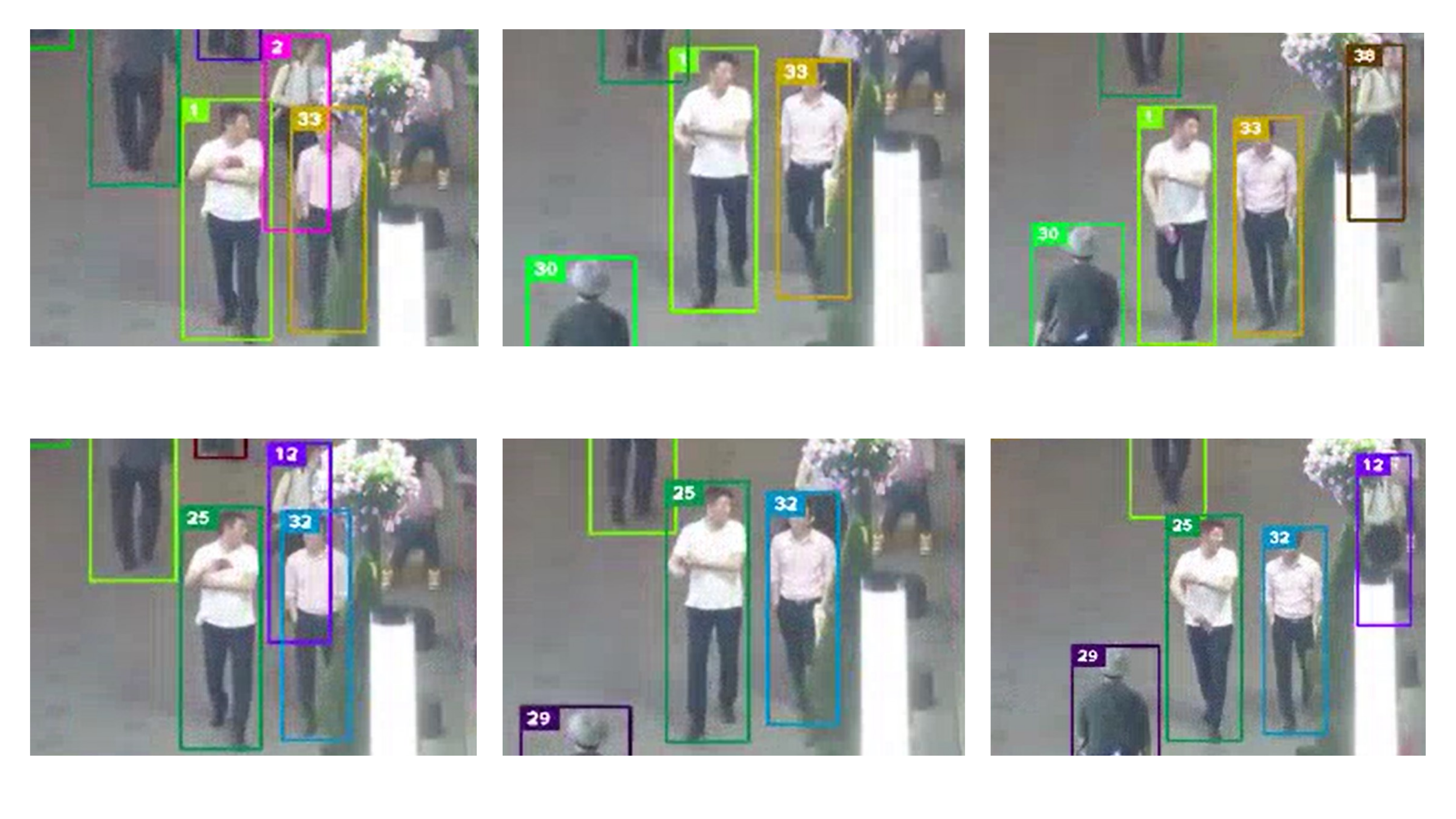}\\
    \label{fig:tracking1}
    \end{minipage}%
    }%
    \subfloat[Two objects exchange their locations.]{
    \begin{minipage}[]{0.5\linewidth}
    \centering
    \includegraphics[width=\linewidth]{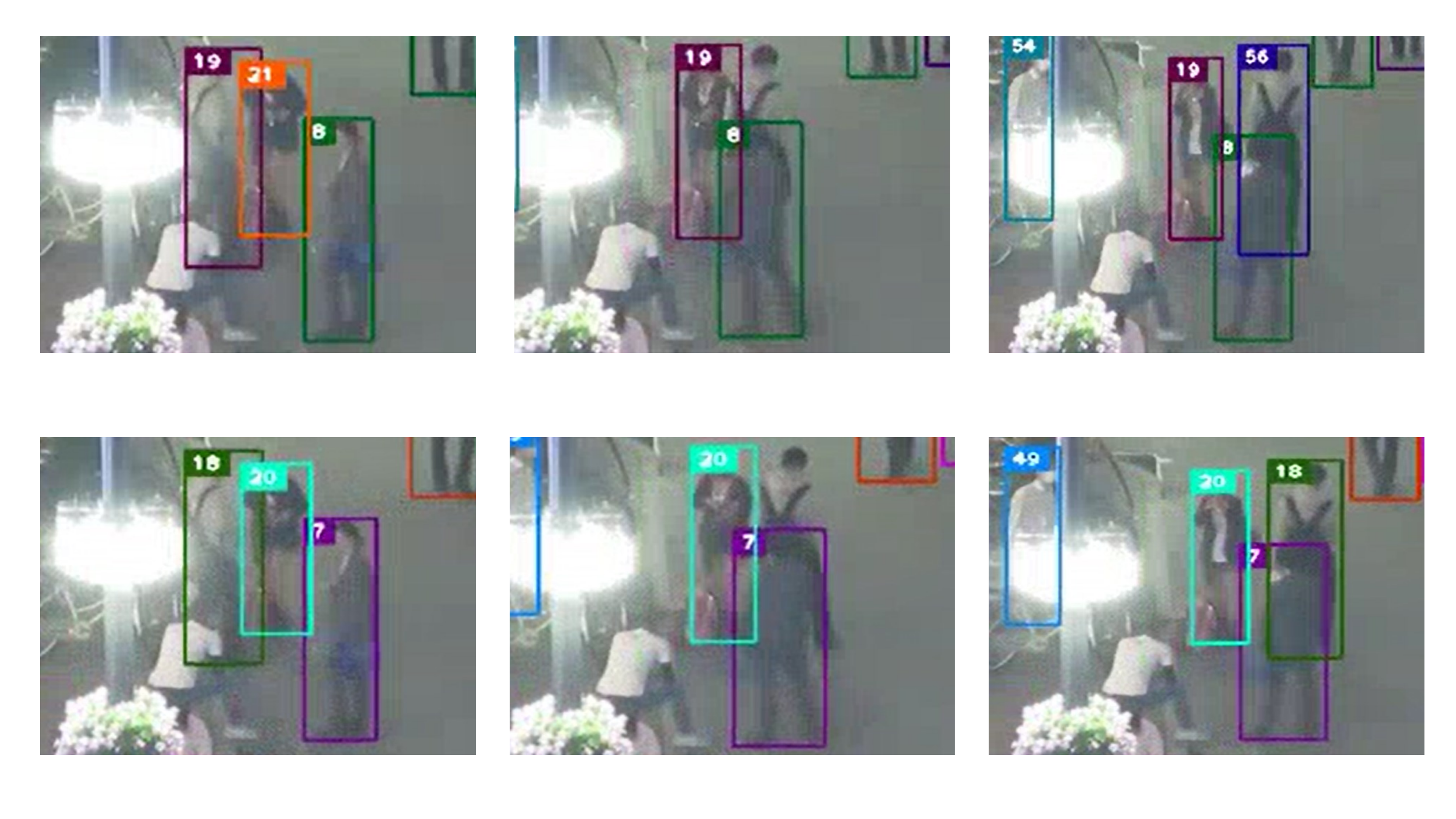}\\
    \label{fig:tracking2}
    \end{minipage}%
    }%
    \caption{Examples of tracking results on MOT17 dataset. The top line is from \emph{DeepSORT}, and the bottom is from \emph{GMTracker}. (a) and (b) are two typical hard cases, in which our method is better than the baseline \emph{DeepSORT}.}
    \label{fig:tracking}
\end{figure*}
\subsection{Inference Time}
We compare the running speed between our new GST algorithm and the original quadratic programming solved by the CVXPY library, as shown in Table~\ref{tbl-speed}. Using the new GST algorithm, with the performance under main metrics almost the same, the speed is about 21$\times$ faster than the original GMTracker, and the solver running time is two orders of magnitude lower than before. Note that the ID switches are much fewer because the matching candidates are filtered before solving the quadratic assignment problem, and the phenomenon of early termination of the tracklet is somewhat alleviated.\par
\begin{table}[t!]
    \center
\resizebox{\columnwidth}{!}{
      \begin{tabular}{l|ll|ccc}
        \toprule
        Methods &FPS* &Solver (s) &IDF1 &MOTA &IDS\\
        \midrule
        GMTracker	&0.987	&8861	&71.5	&62.3	&1017\\
    	  GMTracker+GST &20.7 (21$\times$)	&93.1 (95$\times$)	&71.4	&62.3	&935\\
        \bottomrule
      \end{tabular}}
    \caption{Inference speed comparison between GMTracker and GMTracker+GST on MOT17 \emph{val} set. * Running time of the off-the-shelf object detection algorithm is not considered.}
  \label{tbl-speed}
\end{table}
\begin{table}[h]
    \resizebox{\columnwidth}{!}{
\begin{tabular}{llccccc}
    \toprule
    \multirow{2}{1cm}[-.4em]{Keypoint detector} & \multirow{2}{*}[-.4em]{Matcher}
    & \multicolumn{3}{c}{Pose estimation AUC} & \multirow{2}{*}[-.4em]{P} & \multirow{2}{*}[-.4em]{MS} \\
    \cmidrule(lr){3-5}
    && @5\degree & @10\degree & @20\degree & & \\
    \midrule
    \multirow{7}{1.2cm}[-.4em]{SuperPoint\cite{detone2018superpoint}}
    & NN+mutual & \09.43 & 21.53 & 36.40 & 50.4 & 18.8\\
    & NN+GMS~\cite{bian2017gms} & \08.39 & 18.96 & 31.56 & 50.3 & 19.0\\
    & NN+PointCN~\cite{yi2018learning} & 11.40 & 25.47 & 41.41 & 71.8 & 25.5 \\
    & NN+OANet\cite{zhang2019learning} & 11.76 & 26.90 & 43.85 & 74.0 & 25.7 \\
    & SuperGlue~\cite{sarlin2020superglue} &16.16 &33.81 &51.84 & 84.4 & 31.5 \\
    & SuperGlue*~\cite{sarlin2020superglue} &- &- & 53.38 &- &- \\
    & \b{GMatcher (Ours)} & \b{17.43} & \b{35.75} & \b{54.54} & \b{84.9} & \b{31.6} \\
    \midrule
    \multirow{3}{1.2cm}[-.4em]{Keypoint-free}
&LoFTR~\cite{sun2021loftr} & 22.0  & 40.8 & 57.6 &- &-\\
&LoFTR+QuadTree~\cite{tang2022quadtree} & 24.9  & 44.7 & 61.8&- &- \\
&\textbf{QuadTree+Ours} &\textbf{25.3} &\textbf{45.2} &\textbf{62.4}&- &- \\
    \bottomrule
\end{tabular}}
\caption{Comparisons with SOTA keypoint-based (top) and keypoint-free (bottom) methods on ScanNet. * denotes end-to-end training.}
\label{tbl-sgsota}
\end{table}
\begin{table}[h]
    \begin{center}
      \begin{tabular}{l|cccc}
        \toprule
        Method & PIR&FMR &IR &RR \\
        \midrule
        GeoTransformer~\cite{qin2022geometric} &85.2 &98.0 &70.2 &90.9\\
        \textbf{GeoTransformer~\cite{qin2022geometric}+Ours} &\textbf{86.0} &98.0 &\textbf{70.6} &\textbf{91.7}\\
        \bottomrule
      \end{tabular}
      \end{center}
    \caption{Comparisons between ours and GeoTransformer~\cite{qin2022geometric} on 3DMatch~\cite{zeng20173dmatch} dataset.}
  \label{tab:3dmatch}
  \end{table}

\subsection{Qualitative results on MOT17}
In Fig.~\ref{fig:tracking}, we show the hard cases, in which the baseline tracker \emph{DeepSORT} has ID switches and our \emph{GMTracker} tracks the objects with the right IDs. For example, in Fig.~\ref{fig:tracking1}, DeepSORT fails to track the person with ID-2 and creates a new tracklet ID-38, because the person is occluded by the streetlight and reappears. And in Fig.~\ref{fig:tracking2}, the people with ID-19 and ID-21 exchange their places. DeepSORT can not keep the IDs. The ID-19 drifts to the person with ID-21 and a new ID is assigned to the person with ID-21. However, our GMTracker tracks the objects with right IDs.

\section{Experiments on Image Matching task}
\subsection{Datasets and Evaluation Metrics}
We do the experiments about the Image Matching task on the mainstream indoor camera pose estimation dataset ScanNet~\cite{dai2017scannet}. ScanNet is a large-scale indoor dataset collected using RGB-D cameras. Because it is an indoor dataset, local patterns are always similar in a large area, such as on the white wall and ceramic tile. In image matching task, the general evaluation metrics focus on the camera pose estimation performance, such as the Area Under Curve (AUC) of the pose error with thresholds at $5^{\circ},10^{\circ},20^{\circ}$. Here, the pose error takes the maximum angle error of the rotation matrices and the transformation vectors. Besides, the match precision and the matching score are calculated, mainly reflecting the keypoint matching performance.
\subsection{Implementation Details}
 We take state-of-the-art keypoint-based image matching pipeline SuperGlue~\cite{sarlin2020superglue} as our baseline. Our implementation uses PyTorch~\cite{paszke2019pytorch} framework, and we train our model on 24 NVIDIA RTX 2080Ti GPUs. Adam~\cite{kingma2014adam} optimizer is applied, and the base learning rate is $4\times10^{-5}$ with Cosine Annealing as the learning rate scheduler. Like SuperGlue, we only sample the image pairs with an overlap in $[0.4, 0.8]$ for training. The images and the depth maps are resized to the resolution of $640\times480$. The batch size is 24 and we train the model for 1.2M iterations. In the training stage, only 200 pairs are sampled randomly for each scene in an epoch. During the inference stage, the mutual constraint and the matching score threshold are adopted to filter the mismatch, and the threshold is set to 0.2.
Other experiment settings not written out here are the same as SuperGlue. 
\subsection{Comparisons with SOTA keypoint-based methods}
In Table~\ref{tbl-sgsota}, we show the comparison with other SOTA methods using the same SuperPoint~\cite{detone2018superpoint} keypoint detector on ScanNet test set, which contains 1500 image pairs from \cite{sarlin2020superglue}. Note that we use about half training data and training iterations to outperform SOTA method SuperGlue~\cite{sarlin2020superglue} by 1 pose estimation AUC.\par
We also compare with keypoint-free methods. We regard LoFTR~\cite{sun2021loftr} as our baseline. We add the graph matching module to the coarse matching stage. We replace The attention modules with the quadtree attention modules~\cite{tang2022quadtree}. The results are also shown in Table~\ref{tbl-sgsota}. Compared with LoFTR and quadtree attention baselines, we also have about 0.5 AUC improvement thanks to our learnable graph matching module.
\section{Experiments on Point Cloud Registration task}
To verify the universality of our learnable graph matching method for data association, we also conduct experiments on the Point Cloud Registration task. We choose the baseline as GeoTransformer~\cite{qin2022geometric}. We only change the `Superpoint Matching Module' into our graph matching-based module, i.e., we add the Geometric Transformer module on both vertices and edges instead of only vertices in the original GeoTransformer. We conduct the experiments on the popular 3DMatch~\cite{zeng20173dmatch} dataset. The results are shown in Table~\ref{tab:3dmatch}. We report GeoTransformer results using the official code. The transformation is computed with the local-to-global registration (LGR) method.

\section{Conclusion}

In this paper, we propose a novel learnable graph matching method for data association. Our graph matching method focuses on the pairwise relationship within the view. To make the graph matching module end-to-end differentiable, we relax the QAP formulation into a convex QP and build a differentiable graph matching layer in our Graph Matching Network. We apply our method to the Multiple Object Tracking task, called GMTracker. Taking the second-order edge-to-edge similarity into account, our tracker is more accurate and robust in the MOT task. To speed up our algorithm, we design GST to shrink the area of the feasible region. The experiments of the ablation study and comparison with other state-of-the-art methods both show the efficiency and effectiveness of our method. Moreover, for the Image Matching task, we propose the end-to-end learnable graph matching algorithm based on the SOTA method SuperGlue, called GMatcher, which shows our ability to solve data association tasks generally. The experiments show that we only use half training data and training iterations to outperform SuperGlue by about 1 AUC. In the image matching and point cloud registration tasks, compared with newly reported methods, we achieve SOTA/SOTA comparable performance.


\bibliographystyle{IEEEtran}
\bibliography{egbib}

\begin{IEEEbiography}[{\includegraphics[width=1in,height=1.25in,clip,keepaspectratio]{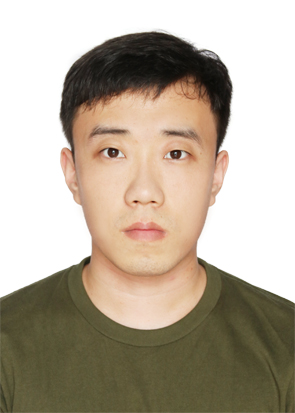}}]{Jiawei He}
  is a PhD student in BRAVE group of Center for Research on Intelligent Perception and Computing (CRIPAC), the National Laboratory of Pattern Recognition (NLPR), Institute of Automation, Chinese Academy of Sciences, Beijing, China, supervised by Prof. Zhaoxiang Zhang. Before this, he got his BS degree in automation from Xi'an Jiaotong University, China, in 2019. His research interests are in Computer Vision, Deep Learning, Learning-based Combinatorial Optimization, including video analysis, graph matching, multiple object tracking, 3D perception, etc.
\end{IEEEbiography}
\begin{IEEEbiography}[{\includegraphics[width=1in,height=1.25in,clip,keepaspectratio]{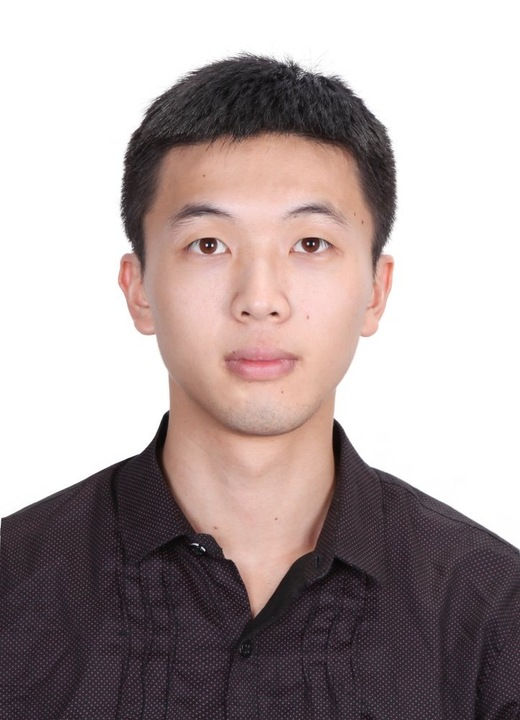}}]{Zehao Huang}
  received the BS degree in automatic control from Beihang University, Beijing,
  China, in 2015. He is currently an algorithm engineer at TuSimple. His research interests include computer vision and image processing.
\end{IEEEbiography}
\begin{IEEEbiography}[{\includegraphics[width=1in,height=1.25in,clip,keepaspectratio]{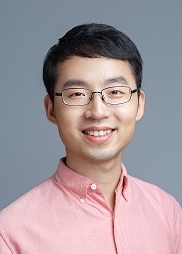}}]{Naiyan Wang}
  is currently the chief scientist of TuSimple. he leads the algorithm research group
  in the Beijing branch. Before this, he got his PhD degree from CSE department, HongKong University of Science and Technology in 2015. His supervisor is Prof. Dit-Yan Yeung. He got his BS degree from Zhejiang University, 2011 under the
  supervision of Prof. Zhihua Zhang. His research interest focuses on applying statistical computational model to real problems in computer vision
  and data mining. Currently, He mainly works on the vision based perception and localization part of autonomous driving. Especially He integrates and improves the cutting-edge technologies in academia, and makes them work properly in the autonomous truck.
\end{IEEEbiography}
\begin{IEEEbiography}[{\includegraphics[width=1in,height=1.25in,clip,keepaspectratio]{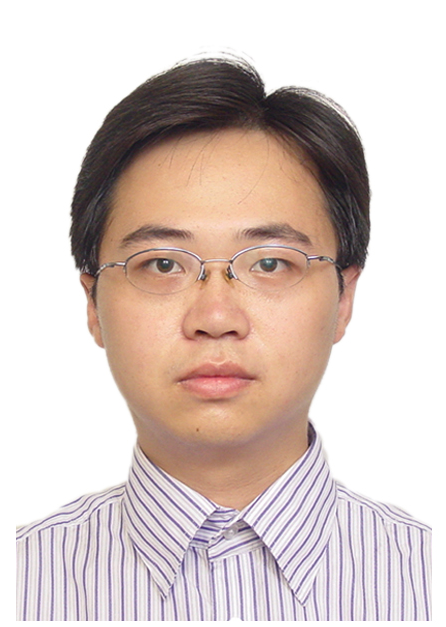}}]{Zhaoxiang Zhang}
received the bachelor's degree in circuits and systems
  from the University of Science and Technology
  of China (USTC) in 2004 and the Ph.D. degree
  from the National Laboratory of Pattern Recognition
  (NLPR), Institute of Automation, Chinese Academy of Sciences (CASIA), Beijing, China, in 2009.
  In October 2009, he joined the School of Computer
  Science and Engineering, Beihang University, and
  worked as an Assistant Professor from 2009 to 2011,
  an Associate Professor from 2012 to 2015, and
  the Vice-Director of the Department of Computer Application Technology
  from 2014 to 2015. In July 2015, he returned to the CASIA, to join as a
  Professor, where he is currently a Professor with the Center for Research on
  Intelligent Perception and Computing. He has published more than 200 papers
  in reputable conferences and journals. His major research interests include
  pattern recognition, computer vision, machine learning, and bio-inspired visual
  computing. He has won the best paper awards in several conferences and
  championships in international competitions. He has served as the Area Chair
  and a Senior PC for many international conferences, such as CVPR, ICCV,
  AAAI, and IJCAI. He has served or is serving as an Associate Editor for IEEE
  TRANSACTIONS ON CIRCUITS AND SYSTEMS FOR VIDEO TECHNOLOGY,
  Pattern Recognition, and Neurocomputing.
\end{IEEEbiography}

\end{document}